\documentclass{article}
\PassOptionsToPackage{numbers, compress}{natbib}

  \usepackage[final]{neurips_2019}

\usepackage[utf8]{inputenc} 
\usepackage[T1]{fontenc}    
\usepackage{hyperref}       
\usepackage{xcolor}
\hypersetup{
    colorlinks, 
    linkcolor={red!50!black},
    citecolor={blue!50!black},
    urlcolor={blue!80!black}
}

\usepackage{url}            
\usepackage{booktabs}       
\usepackage{amsfonts}       
\usepackage{nicefrac}       
\usepackage{microtype}      
\usepackage{float}
\usepackage{graphicx}
\usepackage{booktabs}
\usepackage[inline]{enumitem}
\usepackage{amsmath,amsfonts,amssymb,amscd,amsthm,bm,bbm}
\usepackage{cleveref}
\usepackage{algorithm}
\usepackage{algorithmic}
\usepackage{titlesec}
\titlespacing{\section}{0pt}{\parskip}{-\lineskip}
\titlespacing{\subsection}{0pt}{\parskip}{-\lineskip}
\titlespacing{\subsubsection}{0pt}{\parskip}{-\lineskip}
\newtheoremstyle{new}{5pt}{0pt}{\itshape}{}{\bfseries}{.}{1em}{}
\theoremstyle{new}
\newtheorem{Theorem}{Theorem}

\newtheorem{Definition}{Definition}
\newtheorem{Example}{Example}

\usepackage{etoolbox}
\newcommand{\zerodisplayskips}{%
  \setlength{\abovedisplayskip}{2pt}%
  \setlength{\belowdisplayskip}{2pt}%
  \setlength{\abovedisplayshortskip}{2pt}%
  \setlength{\belowdisplayshortskip}{2pt}}
\appto{\normalsize}{\zerodisplayskips}
\appto{\small}{\zerodisplayskips}
\appto{\footnotesize}{\zerodisplayskips}

\theoremstyle{remark}

\usepackage{wrapfig}
\usepackage{subfig}
\usepackage{xcolor}
\definecolor{airforceblue}{rgb}{0.36, 0.54, 0.66}

\usepackage{balance}

\title{Copulas as High-Dimensional Generative Models:\\
Vine Copula Autoencoders}

\author{%
Natasa Tagasovska\\
Department of Information Systems\\
HEC Lausanne, Switzerland \\
\texttt{natasa.tagasovska@unil.ch} \\
\And
 Damien Ackerer \\
 Swissquote Bank \\
Gland, Switzerland \\
 \texttt{damien.ackerer@swissquote.ch} \\
 \AND
Thibault Vatter \\
Department of Statistics \\
Columbia University, New York, USA \\
 \texttt{thibault.vatter@columbia.edu} \\
}

\begin{document}

\maketitle

\begin{abstract}
  We introduce the vine copula autoencoder (VCAE), a flexible generative model for high-dimensional distributions built in a straightforward three-step procedure.
  First, an autoencoder (AE) compresses the data into a lower dimensional representation.
Second, the multivariate distribution of the encoded data is estimated with vine copulas. 
Third, a generative model is obtained by combining the estimated distribution with the decoder part of the AE.
As such, the proposed approach can transform any already trained AE into a flexible generative model at a low computational cost.
This is an advantage over existing generative models such as adversarial networks and variational AEs which can be difficult to train and can impose strong assumptions on the latent space.
Experiments on MNIST, Street View House Numbers and Large-Scale CelebFaces Attributes datasets show that VCAEs can achieve competitive results to standard baselines. 
\end{abstract}

\section{Introduction}

Exploiting the statistical structure of high-dimensional distributions behind audio, images, or video data is at the core of machine learning.
Generative models aim not only at creating feature representations, but also at providing means of sampling new realistic data points. 
Two classes are typically distinguished: \emph{explicit} and \emph{implicit} generative models. 
Explicit generative models make distributional assumptions on the data generative process.
For example, \emph{variational autoencoders} (VAEs) assume that the latent features are independent and normally distributed \cite{kingma2013auto:vae}.
Implicit generative models make no statistical assumption but leverage another mechanism to transform noise into realistic data.
For example, \emph{generative adversarial networks} (GANs) use a discriminant model penalizing the loss function of a generative model producing unrealistic data \cite{goodfellow2014generative}.
Interestingly, \emph{adversarial autoencoders} (AAEs) combined both features as they use a discriminant model penalizing the loss function of an encoder when the encoded data distribution differs from the prior (Gaussian) distribution \cite{makhzani2015adversarial}.
All of these new types of generative models have achieved unprecedent results and also proved to be computationally more efficient than the first generation of deep generative models which require Markov chain Monte Carlo methods \cite{hinton2006fast,hinton2006reducing}.
However, adversarial approaches require multiple models to be trained,  leading to difficulties and computational burden \cite{radford2015unsupervised, gulrajani2017improved, grnarova2018evaluating}, and variational approaches make (strong) distributional assumptions, potentially detrimental to the generative model performance \cite{rezende2015variational}.

\begin{wrapfigure}{r}{0.5\textwidth}
\centering
\includegraphics[width=0.48\textwidth]{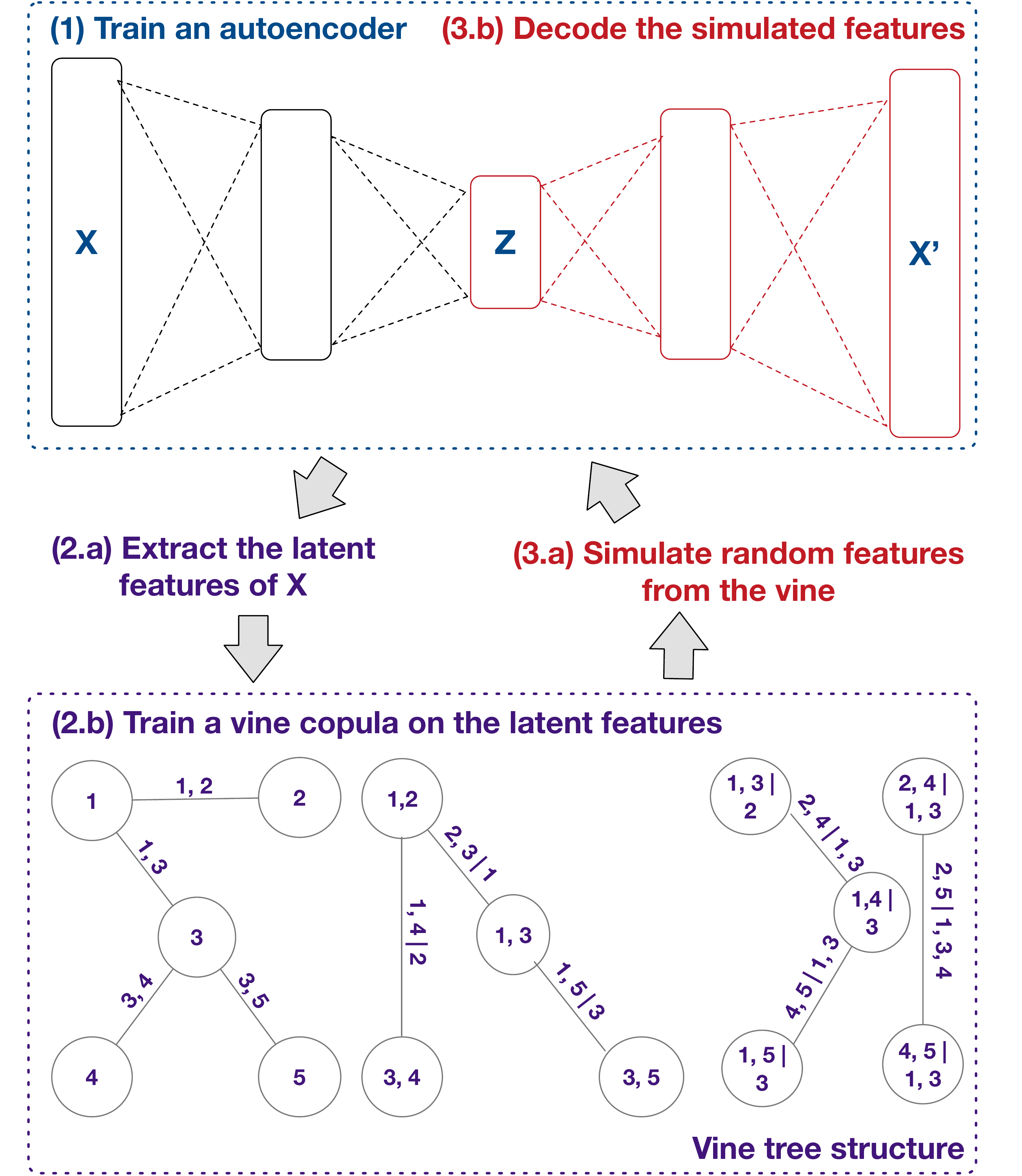}
\caption{Conceptual illustration of a VCAE.}
\label{fig:vcae}
\end{wrapfigure}

We present a novel approach to construct a generative model which is simple, makes no prior distributional assumption (over the input or latent space), and is computationally efficient: the \emph{vine copula autoencoders} (VCAEs).
Our approach, schematized in~\Cref{fig:vcae} combines three tasks. 
First, an autoencoder (AE) is trained to provide high-quality embeddings of the data.
Second, the multivariate distribution of the encoded train data is estimated with vine copulas, namely, a flexible tool to construct high-dimensional multivariate distributions \cite{Bedford01,Bedford02,aasczadofrigessibakken2009}.
Third, a generative model is obtained by combining the estimated vine copula distribution with the decoder part of the AE.

In other words, new data is produced by decoding random samples generated from the vine copula.
An already trained AE can thus be transformed into a generative model, where the only additional cost would be the estimation of the vine copula.
We show in multiple experiments that this approach performs well in building generative models for the MNIST, Large-Scale CelebFaces Attributes, and Street View House Numbers datasets.
To the best of our knowledge, this is the first time that vine copulas are used to construct generative models for very high dimensional data (such as images).

Next, we review the related work most relevant to our setting.
The most widespread generative models nowadays focus on synthetic image generation, and mainly fall into the GAN or VAE categories, some interesting recent developments include \cite{metz2016unrolled,chen2016infogan, gulrajani2017improved, tolstikhin2017wasserstein, higgins2016beta, chavdarova2018sgan, bouchacourt2018multi}. 
These modern approaches have been largely inspired by previous generative models such as belief networks \cite{hinton2006fast}, independent component analysis \cite{hyvarinen2000independent} or denoising AEs \cite{vincent2008extracting}.
Part of their success can be attributed to the powerful neural network architectures which provide high quality feature representations, often using Convolutional architectures \cite{lecun1995convolutional}.
A completely different framework to model multivariate distributions has been developed in the statistical literature: the so-called \emph{copulas}. 
Thanks to their ability to capture complex dependence structures, copulas have been applied to a wide range of scientific problems, and their successes have led to continual advances in both theory and open-source software availability.
We refer to \cite{nelsen2007introduction,joe2014dependence} for textbook introductions. 
More recently, copulas also made their way into machine learning research \cite{Liu2009,Elidan2013,LopezPaz2013,Tran2015,Lopez-Paz2016,Chang2016, tagasovska2018nonparametric, kulkarni2018generative}.
However, copulas have not yet been employed in constructing high dimensional generative models.
While \cite{Li2014,Patki2016a} use copulas for synthetic data generation, they rely on strong parametric assumptions.
In this work, we illustrate how \emph{nonparametric vine copulas} allow for arbitrary density estimation \cite{Nagler2016}, which in turn can be used to sample realistic synthetic datasets.

Because their training is relatively straightforward, VCAEs have some advantages over GANs.
For instance, GANs require some complex modifications of the baseline algorithm in order to avoid mode collapse, whereas vines naturally fit multimodal data.
Additionally, while GANs suffer from the ``exploding gradients'' phenomenon (e.g., see \cite{grnarova2018evaluating}) and require careful monitoring of the training and early stopping, this is not an issue with VCAEs as they are built upon standard AEs.


To summarize, the contribution of this work is introducing a novel, competitive generative model based on copulas and AEs.
There are three main advantages of the proposed approach.
First, it offers modeling flexibility by avoiding most distributional assumptions.
Second, training and sampling procedures for high-dimensional data are straightforward.
Third, it can be used as a plug-in allowing to turn any AE into generative model,
simultaneously allowing it to serve other purposes (e.g., denoising, clustering).

The remainder of the paper is as follows.
\Cref{sec:vines} reviews vine copulas as well as their estimation and simulation algorithms.
\Cref{sec:vine_ae} discusses the VCAE approach.
\Cref{sec:expe} presents the results of our experiments.
\Cref{sec:ccl} concludes and discusses future research.
The supplementary material contains further information on algorithm and experiments, as well as additional experiments.

\section{Vine copulas}
\label{sec:vines}

\subsection{Preliminaries and motivation}
A \emph{copula}, from the latin word \emph{link}, flexibly ``couples'' marginal distributions into a joint distribution.
As such, copulas allow to construct joint distributions with the same margins but different dependence structures, or conversely by fixing the dependence structure and changing the individual behaviors.
Thanks to this versatility, there has been an exponentially increasing interest in copula-based models over the last two decades. 
One important reason lies in the following theorem.

\begin{Theorem}[Sklar's theorem \cite{sklar1959}]\label{thm:sklar}
  The continuous random vector $\bm X = (X_1, \dots, X_d)$ has joint distribution $F$ and marginal distributions $F_1, \dots, F_d$ if and only if there exist a unique copula \footnote{A copula is a distribution function with uniform margins.} $C$, which is the joint distribution of $\bm U = (U_1, \dots, U_d) = \bigl(F_1(X_1), \dots, F_d(X_d)\bigr)$. 
\end{Theorem}

Assuming that all densities exist, we can write $f(x_1, \dots, x_d) = c\bigl\{ u_1, \dots, u_d \bigr\} \times \prod_{k=1}^d f_k(x_k)$,
where $u_i = F_i(x_i)$ and $f,c, f_1, \dots, f_d$ are the densities corresponding to $F, C, F_1, \dots, F_d$ respectively. 
As such, copulas allow to decompose a joint density into a product between the marginal densities $f_i$ and the dependence structure represented by the copula density $c$.

This has an important implication for the estimation and sampling of copula-based marginal distributions: algorithms can generally be built into two steps.
For instance, estimation is often done by estimating the marginal distributions first, 
and then using the estimated distributions to construct pseudo-observations via the probability integral transform before estimating the copula density.
Similarly, synthetic samples can be obtained by sampling from the copula density first, and then using the inverse probability integral transform to transform the copula sample back to the natural scale of the data.
We give a detailed visual example of both the estimation and sampling of (bivariate) copula-based distributions in \Cref{fig:sampling}.
We also refer to \Cref{sec:copula} or the textbooks \cite{nelsen2007introduction} and \cite{joe2014dependence} for more detailed introductions on copulas.

The availability of higher-dimensional models is rather limited, yet there exists numerous parametric families in the bivariate case. 
This has inspired the development of hierarchical models, constructed from cascades of bivariate building blocks:
the \emph{pair-copula constructions} (PCCs), also called \emph{vine copulas}.
Thanks to its flexibility and computational efficiency, this new class of simple yet versatile models
has quickly become a hot-topic of multivariate analysis \cite{Aas2016}.  

\subsection{Vine copulas construction}
\label{sec:vine_const}

Popularized in \cite{Bedford01, Bedford02, aasczadofrigessibakken2009}, PCCs model the joint distribution of a random vector by decomposing the problem into modeling pairs of conditional random variables, making the construction of complex dependencies both flexible and yet tractable.
Let us exemplify such constructions using a three dimensional vector of continuously distributed random variables $X = (X_1, X_2, X_3)$. 
The joint density $f$ of $X$ can be decomposed as
\begin{align}
\label{eq:3d_vine}
	f  &= f_1\, f_2\, f_3\, c_{1,2}\, c_{2,3}\, c_{1,3| 2},
\end{align}
where we omitted the arguments for the sake of clarity, 
and $f_1, f_2, f_3$ are the marginal densities of $X_1, X_2, X_3$, 
$c_{1, 2}$ and $c_{2, 3}$ are the joint densities of $(F_1(X_1),F_2(X_2))$ and $(F_2(X_2),F_3(X_3))$, 

$c_{1, 3| 2}$ is the joint density of $(F_{1|2} (X_1 | X_2),F_{3|2} (X_3|X_2)) | X_2$.

The above decomposition can be generalized to an arbitrary dimension $d$ and leads to tractable and flexible probabilistic models  \cite{Joe97, Bedford01, Bedford02}.
While a decomposition is not unique, it can be organized as a graphical model, a sequence of $d - 1$ nested trees, called \emph{regular vine}, \emph{R-vine}, or simply \emph{vine}.
Denoting $T_m = (V_m, E_m)$ with $V_m$ and $E_m$ the set of nodes and edges of tree $m$ for $m = 1, \dots, d-1$, the sequence is a vine if it satisfies a set of conditions guaranteeing that the decomposition leads to a \emph{valid joint density}. 
The corresponding tree sequence is then called the \emph{structure} of the PCC and has important implications to design efficient algorithms for the estimation and sampling of such models (see \Cref{sec:estimation} and \Cref{sec:sampling}).

Each edge $e$ is associated to a bivariate copula $c_{j_e, k_e | D_e}$ (a so-called \emph{pair-copula}), with the set $D_e \in \left\{1, \cdots, d \right\}$ and the indices $j_e, k_e  \in \left\{1, \cdots, d \right\}$ forming respectively its \emph{conditioning set} and the \emph{conditioned set}. 
Finally, the joint copula density can be written as the product of all pair-copula densities
$c = \prod_{m=1}^{d-1} \prod_{e \in E_m} c_{j_e, k_e | D_e}$. 
In the following two sections, we discuss two topics that are important for the application of vines as generative models: estimation and simulation.
For further details, we refer to the numerous books and surveys written about them \cite{Czado2010,Kurowicka2010,stoeber2012,czado2013,Aas2016}, as well as \Cref{sec:vines_app}.

\subsection{Sequential estimation}
\label{sec:estimation}
To estimate vine copulas, it is common to follow a sequential approach \cite{aasczadofrigessibakken2009,Haff2013,Nagler2016}, which we outline below. 
Assuming that the vine structure is known, the pair-copulas of the first tree, $T_1$, can be directly estimated from the data.
But this is not as straightforward for the other trees, since data from the densities $c_{j_e, d_e |D_e}$ are not observed.
However, it is possible to sequentially construct ``pseudo-observations'' using appropriate data transformations, leading to the following estimation procedure, starting with tree $T_1$: for each edge in the tree, estimate all pairs, construct pseudo-observations for the next tree, and iterate.
The fact that the tree sequence $T_1, T_2, \dots,T_{d-1}$ is a regular vine guarantees that at any step in this procedure, all required pseudo-observations are available.
Additionally to \Cref{sec:vines_est} and \Cref{sec:vines_est2}, we further refer to \cite{aasczadofrigessibakken2009,Brechmann2012,czado2013,Dissmann2013,Brechmann2015,killiches2017}  for model selection methods and to \cite{czado2012,stoeber2013,brechmann2013c,Haff2013,Schepsmeier2014} for more details on the inference and computational challenges related to PCCs. 
Importantly, vines can be truncated after a given number of trees \cite{Brechmann2012,Brechmann2014a,Brechmann2015a} by setting pair-copulas in further trees to independence.
\paragraph{Complexity} Because there are $d$ pair-copulas in $T_1$, $d-1$ pair-copulas in $T_2$, $\dots$, and a single pair-copula in $T_{d-1}$, the complexity of this algorithm is $O(f(n) \times d \times \mbox{truncation level})$, where $f(n)$ is the complexity of estimating a single pair and the truncation level is at most $d - 1$.
In our implementation, described \Cref{sec:implementation}, $f(n) = O(n)$.
  

\subsection{Simulation}
\label{sec:sampling}
Additionally to their flexibility, vines are easy to sample from using inverse transform sampling.
Let $C$ be a copula and $U = (U_1, \dots, U_d) $ is a vector of independent $U(0,1)$ random variables. 
Then, define $V = (V_1, \dots, V_d)$ through $V_1 = C^{-1}(U_1) $, $V_{2} = C^{-1}(U_2|U_1) $, and so on until $V_d =C^{-1}(U_d|U_1,\dots,U_{d-1})$, with $C(v_k|v_1, \dots,v_{k-1})$ is the conditional distribution of $V_k$ given $V_1, \dots, V_{k-1}$, $k = 2,\dots,d$.
In other words, $V$ is the inverse Rosenblatt transform \cite{rosenblatt1952} of $U$.
It is then straightforward to notice that $V \sim C$, which can be used to simulate from $C$. 
As for the sequential estimation procedure, it turns out that
\begin{itemize}
  \item  the fact that the tree sequence $T_1, T_2, \dots,T_{d-1}$ is a vine guarantees that all the required conditional bivariate copulas are available (see Algorithm 2.2 of \cite{Dissmann2013}),
  \item the complexity of the algorithm $O(n \times d \times \mbox{truncation level})$, since $f(n)$ is trivially the complexity required for one inversion multiplied by the number of generated samples.
\end{itemize}
Furthermore, there exist analytical expressions or good numerical approximations of such inverses for common parametric copula families.
We refer to \Cref{sec:implementation} for a discussion of the inverse computations for nonparametric estimators.

\subsection{Implementation}
\label{sec:implementation}

\begin{figure}[H]
\centering
\includegraphics[width=0.95\textwidth]{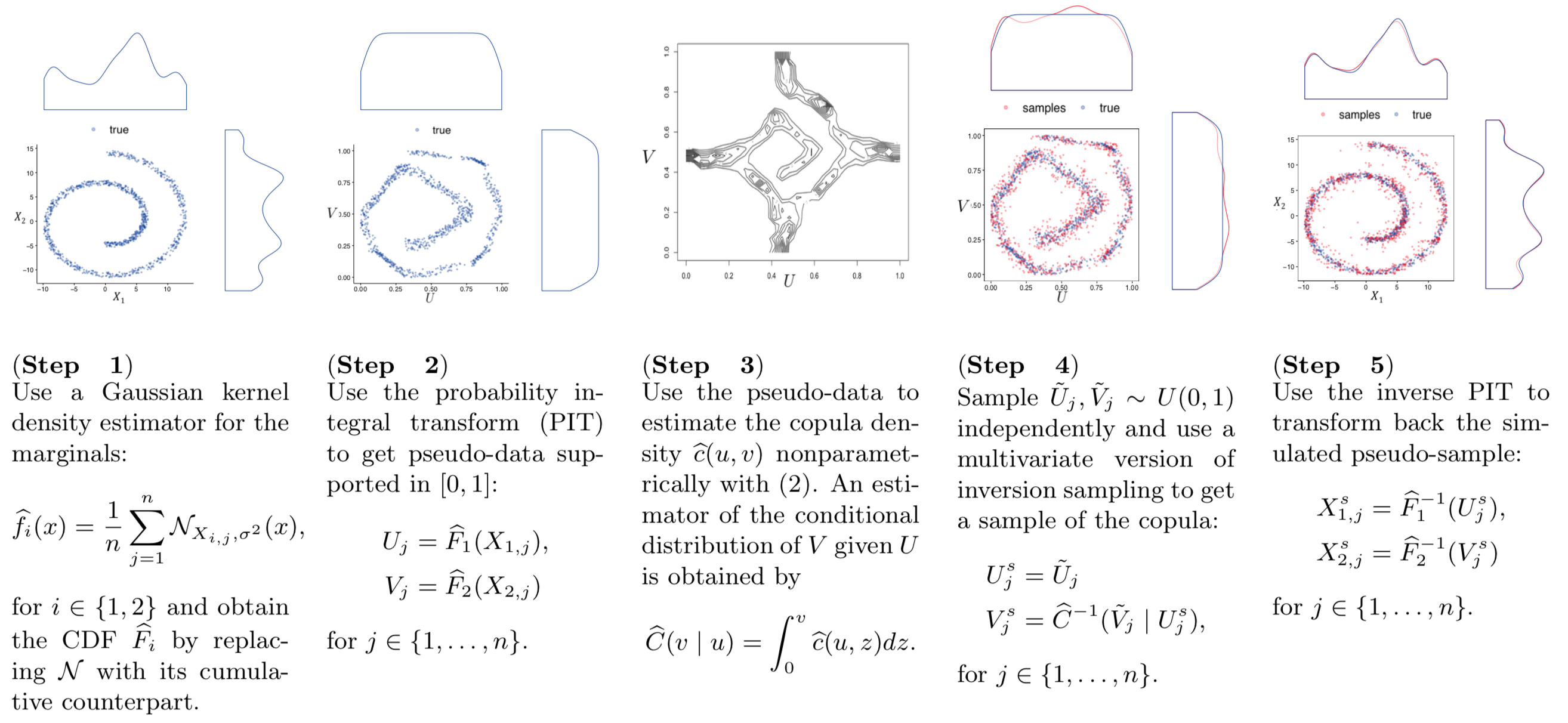}
\caption{Estimation and sampling algorithm for a pair copula.}
\label{fig:sampling}
\end{figure}

To avoid specifying the marginal distributions, we estimate them using a Gaussian kernel with a bandwidth chosen using the direct plug-in methodology of \cite{Sheather1991}.
The observations can then be mapped to the unit square using the probability integral transform (PIT).
See steps 1 and 2 of \Cref{fig:sampling} for an example.

Regarding the copula families used as building blocks for the vine, one can contrast parametric and nonparametric approaches.
As is common in machine learning and statistics, the default choice is the Gaussian copula.
In \Cref{sec:motiv}, we show empirically why this assumption (allowing for dependence between the variables but still in the Gaussian setting) can be too simplistic, resulting in failure to deliver even for three dimensional datasets.

Alternatively, using a nonparametric bivariate copula estimator provides the required flexibility.
However, the bivariate Gaussian kernel estimator, targeted at densities of unbounded support, cannot be directly applied to pair-copulas, which are supported in the unit square.
To get around this issue, the trick is to transform the data to standard normal margins before using a bivariate Gaussian kernel.
Bivariate copulas are thus estimated nonparametrically using the transformation estimator \cite{scaillet2007,LopezPaz2013,Nagler2016,Geenens2017} defined as
\begin{align}\label{eq:tll}
\widehat{c}(u,v)=
\frac{1}{n}\sum_{j=1}^n \frac{\mathcal{N}(\Phi^{-1}(u), \Phi^{-1}(v) | \Phi^{-1}(u_j), \Phi^{-1}(v_j),  \Sigma)}{\phi\left( \Phi^{-1}(u) \right) \phi\left( \Phi^{-1}(v) \right)},
\end{align}
where $\mathcal{N}(\cdot, \cdot | \upsilon_1, \upsilon_2, \Sigma)$ is a two-dimensional Gaussian density with mean $\upsilon_1, \upsilon_2,$ and covariance matrix  $\Sigma = n^{-1/3}\, \mbox{Cor}(\Phi^{-1}(U),\Phi^{-1}(V))$.
For the notation we let $\phi, \Phi$ and $\Phi^{-1} $ to be the standard Gaussian density, distribution and quantile function respectively.
See step 3 of \Cref{fig:sampling} for an example.

Along with vines-related functions (i.e., for sequential estimation and simulation), the Gaussian copula and \eqref{eq:tll} are implemented in C++ as part of \texttt{vinecopulib} \cite{vinecopulib}, a header-only C++ library for copula models based on \texttt{Eigen} \cite{eigenweb} and \texttt{Boost} \cite{Schaling2011}.
In the following experiments, we use the R interface \cite{R} interface to \texttt{vinecopulib} called \texttt{rvinecopulib} \cite{rvinecopulib}, which also include \texttt{kde1d} \cite{Nagler2018} for univariate density estimation.

Note that inverses of partial derivatives of the copula distribution corresponding to \eqref{eq:tll} are required to sample from a vine, as described in \Cref{sec:sampling}.
Internally, \texttt{vinecopulib} constructs and stores a grid over $[0,1]^2$ along with the evaluated density at the grid points.
Then, bilinear interpolation is used to efficiently compute the copula distribution {\scriptsize$\widehat{C}(u,v)$} 
 and its partial derivatives.
Finally, \texttt{vinecopulib} computes the inverses by numerically inverting the bilinearly interpolated quantities using a vectorized version of the bisection method, and we show a copula sample example as step 4 of \Cref{fig:sampling}.
The consistency and asymptotic normality of this estimator are derived in \cite{Geenens2017} under assumptions described in \Cref{sec:est_assumptions}.

To recover samples on the original scale, the simulated copulas samples, often called pseudo-samples, are then transformed using the inverse PIT, see step 5 of \Cref{fig:sampling}.
In \Cref{sec:toy}, we show that this estimator performs well on two toy bivariate datasets that are typically challenging for GANs: a grid of isotropic Gaussians and the swiss roll.

\subsection{Vines as generative models}
\label{sec:motiv}

To exemplify the use of vines as generative models, let us consider as a running example a three dimensional dataset  $X_1, X_2, X_3$ with $X_1, X_2 \sim U [-5, 5]$ and $X_3 = \sqrt{X_1^2 + X_2^2} + U[-0.1, 0.1]$. 
The joint density can be decomposed as in the right-hand side of \eqref{eq:3d_vine}, and estimated following the procedures described in \Cref{sec:implementation} and \Cref{sec:estimation}. 
With the structure and the estimated pair copulas, we can then use vines as generative models. 

\begin{wrapfigure}{r}{0.6\textwidth}
\centering
\includegraphics[width=0.58\textwidth]{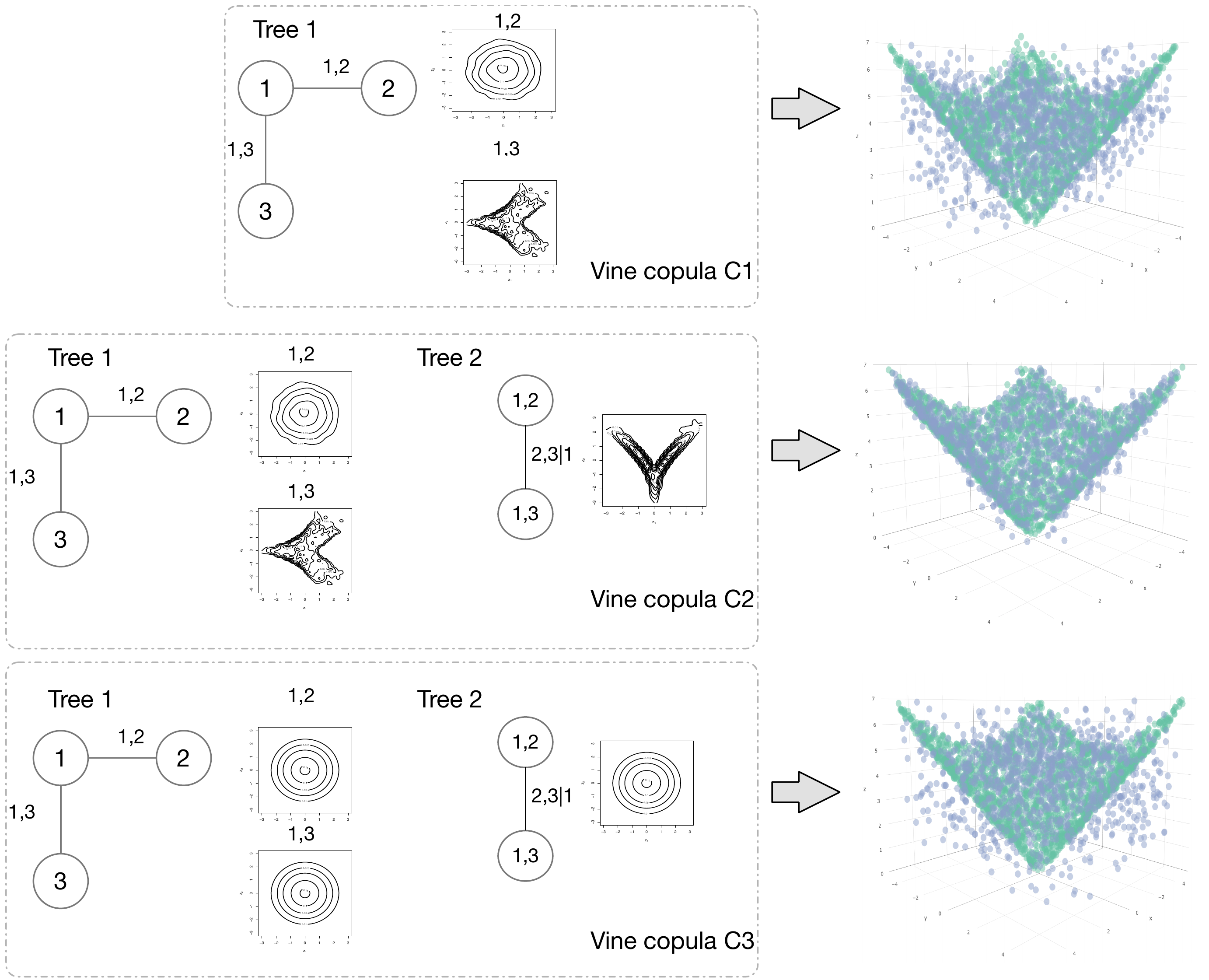}
\caption{Simulation with different truncation levels, top to bottom - 1 level truncated vine, 2 levels non-parametric vine, 2 levels Gaussian vine.}
\label{fig:vine_gen_ex}
\end{wrapfigure}

In \Cref{fig:vine_gen_ex}, we showcase three models.
C1 is a nonparametric vine truncated after the first tree. In other words, it sets $c_{2, 3|1}$ to independence.
C2 is a nonparametric vine with two trees.
C3 is a Gaussian vine with two trees.
On the left panel, we show their vine structure, namely the trees and the pair copulas.
On the right panel, we present synthetic samples from each of the models in blue, with the green data points corresponding to $\sqrt{X_1^2+ X_2^2}$. 

Comparing C1 to C2 allows to understand the truncation effect: 
C2, being more flexible (fitting richer/deeper model), captures better the features of the joint distribution.
It can be deduced from the fact that data generated by C2 looks like uniformly spread around the $\sqrt{X_1^2+ X_2^2}$ surface, while data generated by C1 is spread all around.
It should be noted that, in both cases, the nonparametric estimator captures the fact that $X_1$ and $X_2$ are independent, as can be seen from the contour densities on the left panel.
Regarding C3, it seems clear that Gaussian copulas are not suited to handle this kind of dependencies: for such nonlinearities, the estimated correlations are (close to) zero, as can be seen from the contour densities on the left panel.

With this motivation, the next section is dedicated to extending the vine generative approach to high dimensional data. 
While vines are theoretically suitable for fitting and sampling in high dimensions, they have been only applied to model a few thousands of variables.
The reason is mainly that state-of-the-art implementations were geared towards applications such as climate science and financial risk computations.
While software such a \texttt{vinecopulib} satisfies the requirements of such problems, even low-resolution images (e.g., $64 \times 64 \times 3$) are beyond its current capabilities. 
To address this challenge, we can rely on the embedded representations provided by neural networks.  


\section{Vine copula autoencoders}\label{sec:vine_ae}

The other building block of the VCAE is an \emph{autoencoder} (AE) \cite{bourlard1988auto, hinton1994autoencoders}. 
These neural network models typically consist of two parts: an \emph{encoder} $f$ mapping a datum $X$ from the original space $\mathcal{X}$ to the latent space $\mathcal{Y}$, and a decoder $g$ mapping a latent code $Y$ from the latent space $\mathcal{Y}$ to the original space $\mathcal{X}$. 
The AE is trained to reconstruct the original input with minimal reconstruction loss, that is $X^\prime \approx g(f(X))$. 


However, AEs simply learn the most informative features to minimize the reconstruction loss, and therefore cannot be considered as generative models. 
In other words, since they do not learn the distributional properties of the latent features \cite{bengio2013representation}, they cannot be used to sample new data points. 
Because of the latent manifold's complex geometry, attempts using simple distributions (e.g., Gaussian) for the latent space may not provide satisfactory results.

Nonparametric vines naturally fill this gap. 
After training an AE, we use its encoder component to extract lower dimensional feature representations of the data. 
Then, we fit a vine without additional restrictions on the latent distribution.
With this simple step, we transform AEs into generators, by systematically sampling data from the vine copula, following the procedure from~\Cref{sec:sampling}. 
Finally, we use the decoder to transform the samples from vine in latent space into simulated images in pixel space.
A schematic representation of this idea is given in \Cref{fig:vcae} and pseudo-code for the VCAE algorithm can be found in \Cref{sec:vcae_algo}.

The vine copula is fitted post-hoc for two reasons.
First, since the nonparametric estimator is consistent for (almost) any distribution, the only purpose of the AE is to minimize the reconstruction error.
The AE's latent space is unconstrained and the same AE can be used for both conditional and unconditional sampling.
Second, it is unclear how to train a model that includes a nonparametric estimator since it has no parameters, there is no loss function to minimize or gradients to propagate.
One possibility would be using spline estimators, which would allow to train the model end-to-end by fitting the basis expansion's coefficients.
However, spline estimators of copula densities have been empirically shown to have inferior performance than the transformation kernel estimator~\cite{nagler2017nonparametric}.

There is some leeway in modeling choices related to the vine. 
For instance, the number of trees as well as the choice of copula family (i.e., Gaussian or nonparametric) have an impact of the synthetic samples, as sharper details are expected from more flexible models.
Note that one can adjust the characteristics of the vine until an acceptable fit of the latent features even after the AE is trained.

%

\section{Experiments} \label{sec:expe}

To evaluate VCAEs as generative models, we follow an experimental setup similar as related works on GANs and VAEs.
We compare vanilla VAEs to VCAEs using the same architectures, but replacing the variational part of the VAEs by vines to obtain the VCAEs.
From the generative adversarial framework, we compare to DCGAN \cite{radford2015unsupervised}.
The architectures for all networks are described in \Cref{sec:architectures}.

Additionally, we explore two modifications of VCAE, (i) Conditional VCAE, that is sampling from a mixture obtained by fitting one vine per class label, and (ii) DEC-VCAE, namely adding a clustering-related penalty as in \cite{xie2016unsupervised}. 
The rationale behind the clustering penalty was to better disentangle the features in the latent space.
In other words, we obtain latent representations where the different clusters (i.e., classes) are better separated, thereby facilitating their modeling.

\subsection{Experimental setup}

\textbf{Datasets and metrics}

We explore three real-world datasets: two small scale - MNIST \cite{lecun-mnisthandwrittendigit-2010} and Street View House Numbers (SVNH) \cite{netzer2011reading}, and one large scale - CelebA \cite{liu2015faceattributes}.
While it is generally common to evaluate models by comparing their log-likelihood on a test dataset, this criterion is known to be unsuitable to evaluate the quality of sampled images \cite{Theis2015}. 
As a result, we use an evaluation framework recently developed for GANs \cite{xu2018empirical}. 
According to \cite{xu2018empirical}, the most robust metrics for two sample testing are the \emph{classifier two sample test} (C2ST, \cite{lopez2016revisiting}) and \emph{mean maximum discrepancy} score (MMD, \cite{gretton2007kernel}).
Furthermore, \cite{xu2018empirical} proposes to use these metrics not only in the pixel space, but over feature mappings in convolution space.
Hence, we also compare generative models in terms of Wasserstein distance, MMD score and C2ST accuracy over ResNet-34 features.
Additionally, we also use the \emph{common inception score} \cite{salimans2016improved} and \emph{Fr\'echet inception distance} (FID, \cite{heusel2017gans}). 
For all metrics, lower values are better, except for inception.
We refer the reader to \cite{xu2018empirical} for further details on the metrics and the implementation.

\textbf{Architectures, hyperparameters, and hardware}

For all models, we fix the AE's architecture as described in \Cref{sec:architectures}.
Parameters of the optimizers and other hyperparameters are fixed as follows.
Unless stated otherwise, all experiments were run with nonparametric vines and truncated after 5 trees.
We use deep CNN models for the AEs in all baselines and follow closely DCGAN \cite{radford2015unsupervised} with batch normalization layers for natural image datasets.
For all AE-based methods, we use the Adam optimizer with learning rate 0.005 and weight decay 0.001 for all the natural image experiments, and 0.001 for both parameters on MNIST.
For DCGAN, we use the recommended learning rate 0.0002 and $\beta_1 = 0.5$ for Adam.
The size of the latent spaces $z$ was selected depending on the dataset's size and complexity. 
For MNIST, we present results with $z=10$, SVHN $z=20$ and for CelebA $z=100$.
We chose to present the values that gave reasonable results for all baselines.
For MNIST, we used batch size of 128, for SVHN 32, and for CelebA batches of 100 samples for training.
All models were trained on a separate train set, and evaluated on hold out test sets of 2000 samples, which is the evaluation size used in \cite{xu2018empirical}.
We used Pythorch 4.1 \cite{paszke2017automatic}, and we provide our code in \Cref{sec:code}.
All experiments were executed on an AWS instance \emph{p2.xlarge} with an NVIDIA K80 GPU, 4 CPUs and 61 GB of RAM.

\subsection{Results}


\textbf{MNIST}

\begin{figure}[h]
\centering
\includegraphics[width=0.75\textwidth]{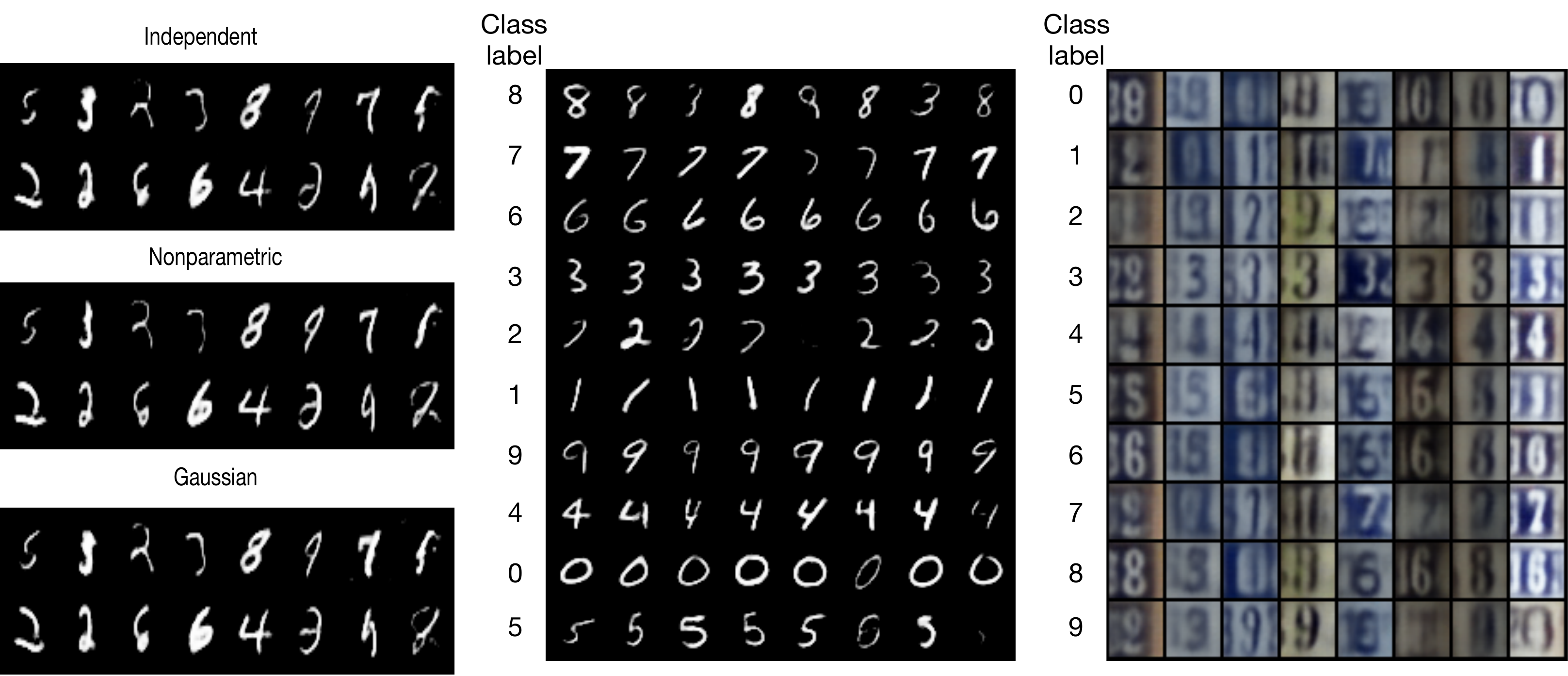}
\caption{Left - impact of copula family selection on \textbf{MNIST}. Middle and Right - random samples of Conditional \emph{VCAE} on \textbf{MNIST} and \textbf{SVHN}.}
\label{fig:interpoation_mnist}
\end{figure}

In \Cref{fig:interpoation_mnist}, we present results from VCAE to understand how different copula families impact the quality of the samples. 
The independence copula corresponds to assuming independence between the latent features as in VAEs.
And the images generated using nonparametric vines seem to improve over the other two.
Within our framework, the training of the AE and the vine fit are independent.
And we can leverage this to perform conditional sampling by fitting a different vine for each class of digit.
We show results of vine samples per digit class in \Cref{fig:interpoation_mnist}.

\textbf{SVHN}

\begin{figure}[t]
\centering
\includegraphics[width=0.9\textwidth]{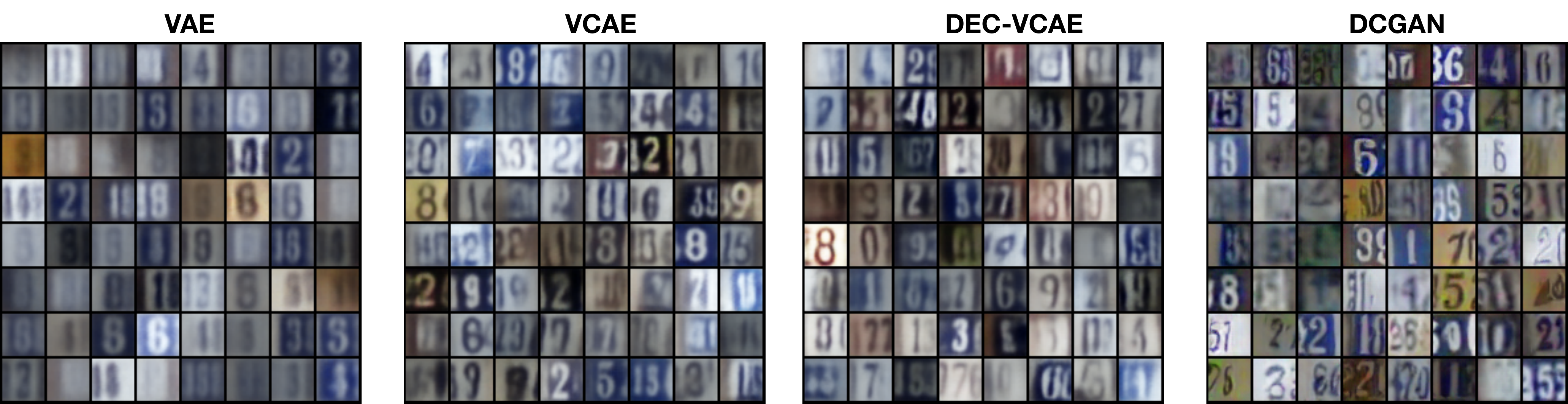}
\caption{Left to right, random samples of \emph{VAE}, \emph{VCAE}, \emph{DEC-VCAE}, and \emph{DCGAN} for \textbf{SVHN}.}
\label{fig:svhn_examples}
\end{figure}

\begin{figure}[h]
\centering
\includegraphics[width=1\textwidth]{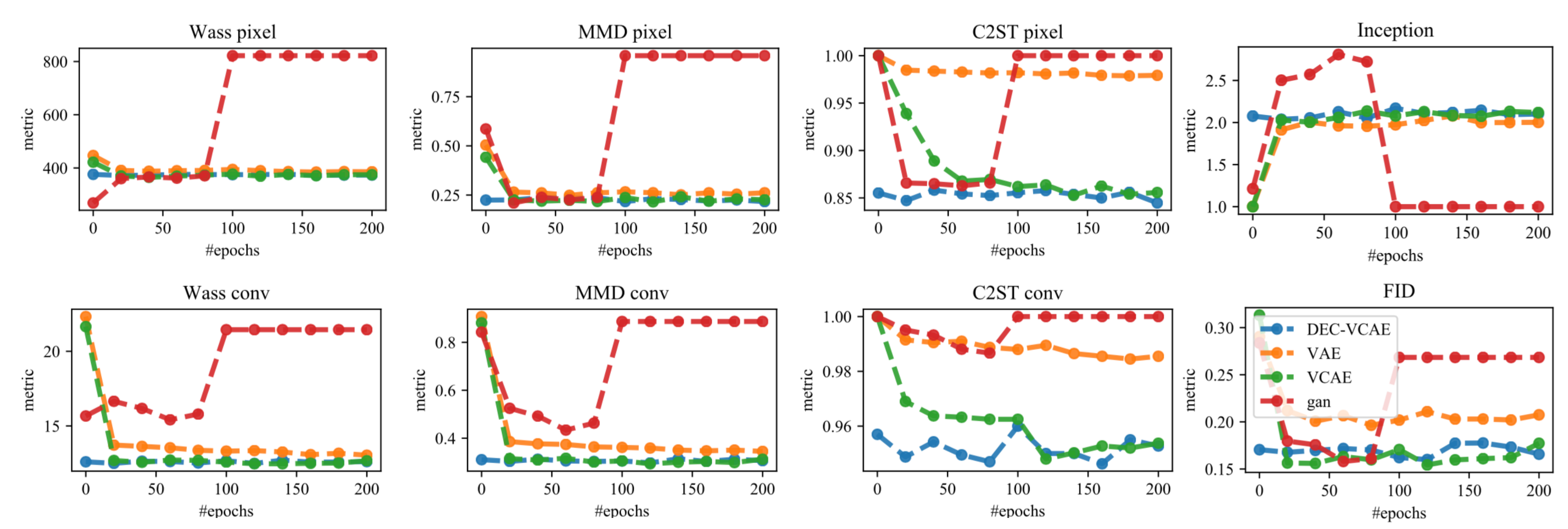}
\caption{Various evaluation scores for all baselines on the \textbf{SVHN} dataset.}
\label{fig:svhn scores}
\end{figure}

The results in \Cref{fig:svhn_examples} show that the variants of vine generative models visually provide sharper images than vanilla VAEs when architectures and training hyper-parameters are the same for all models.
All AE-based methods were trained on latent space $z=20$ for 200 epochs, while for DCGAN we use $z=100$ and evaluate it at its best performance (50 epochs).
In \Cref{fig:svhn scores}, we can see that VCAE and DEC-VCAE have very similar and competitive results to DCGAN (at its best) across all metrics, and both clearly outperform vanila VAE.
Finally, the FID score calculated with regards to $10^4$ real test samples are has $0.205$ for VAE, $0.194$ for DCGAN and $0.167$ for VCAE which shows that VCAE also has slight advantage using this metric.
In \Cref{sec:metrics_mnist_celeba}, \Cref{fig:mnist_scores} and \Cref{fig:celeba_scores} show similar results respectively for the MNIST and CelebA datasets.

\begin{figure}[h]
\centering
\includegraphics[width=0.95\textwidth]{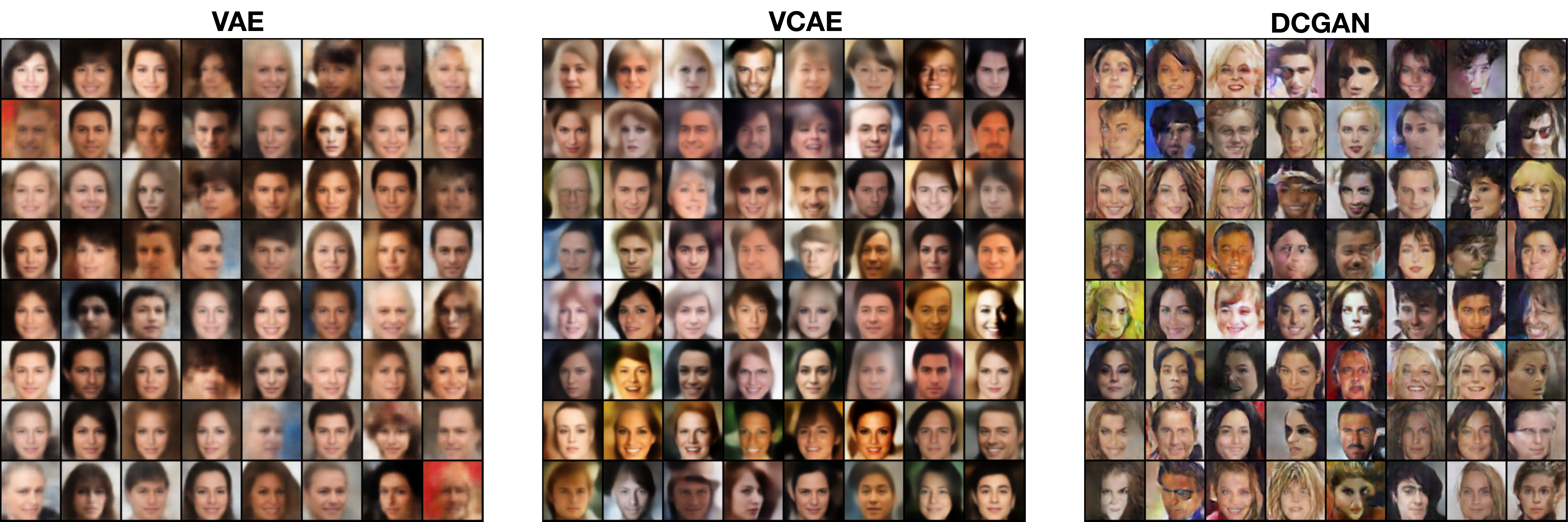}
\caption{Random samples for models trained on the \textbf{CelebA} dataset, for \emph{VAE} and \emph{VCAE} at 200 epochs, and for \emph{DCGAN} best results at 30 epochs.}
\label{fig:vine_ae_celeb}
\end{figure}

\textbf{CelebA}

In the large scale setting, we present results for VCAE, VAE, and DCGAN only, because our GPU ran out of memory on DEC-VCAE.
From the random samples in \Cref{fig:vine_ae_celeb}, we see that, for the same amount of training (in terms of epochs), VCAE results is not only sharper but also produce more diverse samples. 
VAEs improve using additional training, but vine-based solutions achieve better results with less resources and without constraints on the latent space.
Note that, in \Cref{sec:interpolation}, we also study the quality of the latent representation.

\begin{wrapfigure}{r}{0.45\textwidth}
\centering
\includegraphics[width=0.45\textwidth]{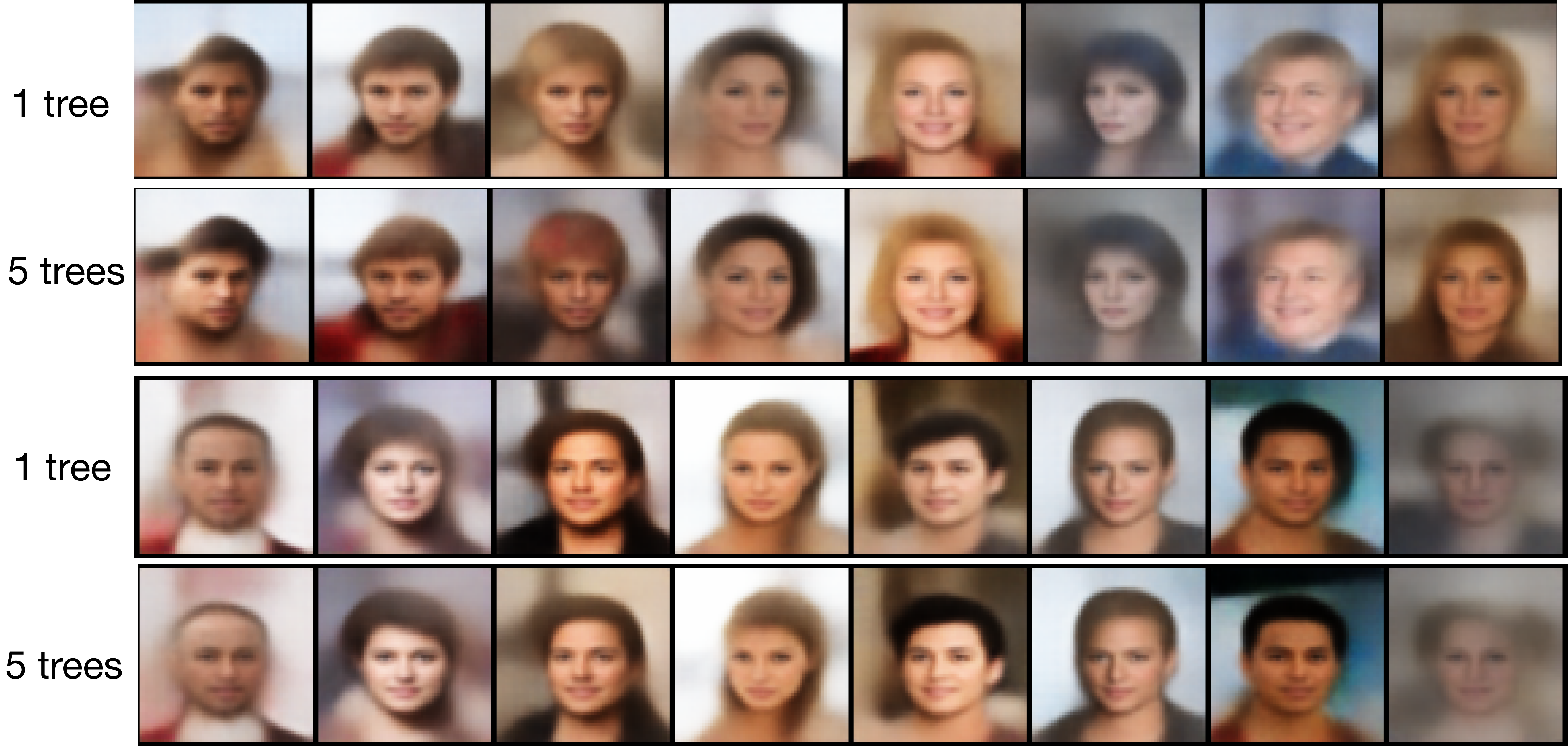}
\caption{Higher truncation - sharper images.\label{fig:celeb_1_vs_5_trees}}
\end{wrapfigure}

To see the effect of the number of trees in the vine structure, we include \Cref{fig:celeb_1_vs_5_trees}, where we can see that from the random sample the vine with five trees provides images with sharper details.
Since, as stated in \Cref{sec:estimation} and \Cref{sec:sampling}, the algorithms complexity increases linearly with the number of trees, we explore the trade-off between computation time and quality of the samples in \Cref{sec:complexity_quality}. 
Results show that, as expected, deeper vines, and hence longer computation times, improve the quality of the generated images.
Finally, as for SVHN, the FID score shows an advantage of the vine-base method over VAEs as we find 0.247 for VAE and 0.233 for VCAE. 
For DCGAN the FID score is 0.169 which is better than VCAE, however, looking at the random batch samples in \Cref{fig:vine_ae_celeb} although GANs outputs sharper images, it is clear that VCAE produces more realistic faces.

\textbf{Execution times}

\begin{wraptable}{r}{0.5\textwidth}
\caption{Execution times.\label{tab:ex_time}}
 \resizebox{0.5\textwidth}{!}{
\begin{tabular}{@{}lccc@{}}
\toprule
& \textbf{\begin{tabular}[c]{@{}c@{}}MNIST\\ (200 epochs)\end{tabular}} & \textbf{\begin{tabular}[c]{@{}c@{}}SVHN\\ (200 epochs)\end{tabular}} & \textbf{\begin{tabular}[c]{@{}c@{}}CelebA\\ (100 epochs)\end{tabular}} \\ \midrule
\textbf{VAE}         & 50 min                                                             & 4h 7 min                                                          & 7h                                                                  \\
\textbf{VCAE}     & 55 min                                                             & 1h 32 min                                                         & 6.5h                                                                \\
\textbf{DEC VCAE} & 101 min                                                            & 2h 35 min                                                         & /                                                                  
\\
\textbf{DCGAN} &              120 min (40 epochs)                                                & 3h 20 min (50 epochs)                                                        & 5h (30 epochs)
 \\ \bottomrule
\end{tabular}}
\end{wraptable}

We conclude the experimental section with \Cref{tab:ex_time} comparing execution times. 
We note that VCAE compares favorably to VAE, which is a ``fair'' observation given that the architectures are alike.
Comparison to DCGAN is more difficult, due to the different nature of the two frameworks (i.e., based respectively on AEs or adversarial).

It should also be noted that the implementation of VCAE is far from optimal for two reasons.
First, we use the \texttt{R} interface to \texttt{vinecopulib} in \texttt{Python} through \texttt{rpy2}.
As such, there is a communication overhead resulting from switching between \texttt{R} and \texttt{Python}.
Second, while \texttt{vinecopulib} uses native \texttt{C++11} multithreading, it does not run on GPU cores.
From our results, this is not problematic, since the execution times are satisfactory.
But VCAE could be much faster if nonparametric vines were implemented in a tensor-based framework.

\section{Conclusion} \label{sec:ccl}

In this paper, we present vine copula autoencoders (VCAEs), a first attempt at using copulas as high-dimensional generative models.
VCAE leverage the capacities of AEs at providing compressed representations of the data, along with the flexibility of nonparametric vines to model arbitrary probability distributions.
We highlight the versatility and power of vines as generative models in high-dimensional settings with experiments on various real datasets.
VCAEs results show that they are comparable to existing solutions in terms of sample quality, while at the same time providing straightforward training along more control over flexibility at modeling and exploration (tuning truncation level, selection of copula families/parameter values).
Several directions for future work and extensions are being considered.
First, we started to experiments with VAEs having flexible distributional assumptions (i.e., by using a vine on the variational distribution).
Second, we plan on studying hybrid models using adversarial mechanisms.
In related work \cite{kulkarni2018generative}, we have also investigated the method's potential for sampling \emph{sequential data} (artificial mobility trajectories). 
There can also be extensions to text data, or investigating which types of vines synthesize best samples for different data types.

\newpage
\bibliographystyle{plainnat}
\bibliography{neurips_2019.bib}

\begin{thebibliography}{82}
\providecommand{\natexlab}[1]{#1}
\providecommand{\url}[1]{\texttt{#1}}
\expandafter\ifx\csname urlstyle\endcsname\relax
  \providecommand{\doi}[1]{doi: #1}\else
  \providecommand{\doi}{doi: \begingroup \urlstyle{rm}\Url}\fi

\bibitem[Aas et~al.(2009)Aas, Czado, Frigessi, and
  Bakken]{aasczadofrigessibakken2009}
K.~Aas, C.~Czado, A.~Frigessi, and H.~Bakken.
\newblock {Pair-Copula Constructions of Multiple Dependence}.
\newblock \emph{Insurance: Mathematics and Economics}, 44\penalty0
  (2):\penalty0 182--198, 2009.

\bibitem[Aas(2016)]{Aas2016}
Kjersti Aas.
\newblock {Pair-copula constructions for financial applications: A review}.
\newblock \emph{Econometrics}, 4\penalty0 (4):\penalty0 43, October 2016.

\bibitem[Bedford and Cooke(2001)]{Bedford01}
Tim Bedford and Roger~M. Cooke.
\newblock {Probability Density Decomposition for Conditionally Dependent Random
  Variables Modeled by Vines}.
\newblock \emph{Annals of Mathematics and Artificial Intelligence}, 32\penalty0
  (1-4):\penalty0 245--268, 2001.

\bibitem[Bedford and Cooke(2002)]{Bedford02}
Tim Bedford and Roger~M. Cooke.
\newblock {Vines -- A New Graphical Model for Dependent Random Variables}.
\newblock \emph{The Annals of Statistics}, 30\penalty0 (4):\penalty0
  1031--1068, 2002.

\bibitem[Bengio et~al.(2013)Bengio, Courville, and
  Vincent]{bengio2013representation}
Yoshua Bengio, Aaron Courville, and Pascal Vincent.
\newblock Representation learning: A review and new perspectives.
\newblock \emph{IEEE transactions on pattern analysis and machine
  intelligence}, 35\penalty0 (8):\penalty0 1798--1828, 2013.

\bibitem[Bouchacourt et~al.(2018)Bouchacourt, Tomioka, and
  Nowozin]{bouchacourt2018multi}
Diane Bouchacourt, Ryota Tomioka, and Sebastian Nowozin.
\newblock Multi-level variational autoencoder: Learning disentangled
  representations from grouped observations.
\newblock In \emph{AAAI}, 2018.

\bibitem[Bourlard and Kamp(1988)]{bourlard1988auto}
Herv{\'e} Bourlard and Yves Kamp.
\newblock Auto-association by multilayer perceptrons and singular value
  decomposition.
\newblock \emph{Biological cybernetics}, 59\penalty0 (4-5):\penalty0 291--294,
  1988.

\bibitem[Brechmann and Joe(2014)]{Brechmann2014a}
Eike~C. Brechmann and Harry Joe.
\newblock {Parsimonious parameterization of correlation matrices using
  truncated vines and factor analysis}.
\newblock \emph{Computational Statistics and Data Analysis}, 77:\penalty0
  233--251, 2014.

\bibitem[Brechmann and Czado(2015)]{Brechmann2015}
Eike~Christian Brechmann and Claudia Czado.
\newblock {COPAR---multivariate time series modeling using the copula
  autoregressive model}.
\newblock \emph{Applied Stochastic Models in Business and Industry},
  31\penalty0 (4):\penalty0 495--514, 2015.

\bibitem[Brechmann and Joe(2015)]{Brechmann2015a}
Eike~Christian Brechmann and Harry Joe.
\newblock {Truncation of vine copulas using fit indices}.
\newblock \emph{Journal of Multivariate Analysis}, 138:\penalty0 19--33, 2015.

\bibitem[Brechmann and Schepsmeier(2013)]{brechmann2013c}
Eike~Christian Brechmann and Ulf Schepsmeier.
\newblock {Modeling dependence with C-and D-vine copulas: The R-package
  CDVine}.
\newblock \emph{Journal of Statistical Software}, 52\penalty0 (3):\penalty0
  1--27, 2013.

\bibitem[Brechmann et~al.(2012)Brechmann, Czado, and Aas]{Brechmann2012}
Eike~Christian Brechmann, Claudia Czado, and Kjersti Aas.
\newblock {Truncated regular vines in high dimensions with application to
  financial data}.
\newblock \emph{Canadian Journal of Statistics}, 40\penalty0 (1):\penalty0
  68--85, March 2012.

\bibitem[Chang et~al.(2016)Chang, Li, Ding, and Dy]{Chang2016}
Yale Chang, Yi~Li, Adam Ding, and Jennifer Dy.
\newblock {A robust-equitable copula dependence measure for feature selection}.
\newblock \emph{AISTATS}, 2016.

\bibitem[Chavdarova and Fleuret(2018)]{chavdarova2018sgan}
Tatjana Chavdarova and Fran{\c{c}}ois Fleuret.
\newblock Sgan: An alternative training of generative adversarial networks.
\newblock In \emph{CVPR}, 2018.

\bibitem[Chen et~al.(2016)Chen, Duan, Houthooft, Schulman, Sutskever, and
  Abbeel]{chen2016infogan}
Xi~Chen, Yan Duan, Rein Houthooft, John Schulman, Ilya Sutskever, and Pieter
  Abbeel.
\newblock Infogan: Interpretable representation learning by information
  maximizing generative adversarial nets.
\newblock In \emph{Advances in neural information processing systems}, pages
  2172--2180, 2016.

\bibitem[Czado(2010)]{Czado2010}
Claudia Czado.
\newblock {Pair-Copula Constructions of Multivariate Copulas}.
\newblock In Piotr Jaworski, Fabrizio Durante, Wolfgang~Karl H{\"{a}}rdle, and
  Tomasz Rychlik, editors, \emph{Copula Theory and Its Applications}, Lecture
  Notes in Statistics, pages 93--109. Springer Berlin Heidelberg, 2010.

\bibitem[Czado et~al.(2012)Czado, Schepsmeier, and Min]{czado2012}
Claudia Czado, Ulf Schepsmeier, and Aleksey Min.
\newblock {Maximum likelihood estimation of mixed C-vines with application to
  exchange rates}.
\newblock \emph{Statistical Modelling}, 12\penalty0 (3):\penalty0 229--255,
  2012.

\bibitem[Czado et~al.(2013)Czado, Brechmann, and Gruber]{czado2013}
Claudia Czado, Eike~Christian Brechmann, and Lutz Gruber.
\newblock {Selection of Vine Copulas}.
\newblock In Piotr Jaworski, Fabrizio Durante, and Wolfgang~Karl H{\"{a}}rdle,
  editors, \emph{Copulae in Mathematical and Quantitative Finance: Proceedings
  of the Workshop Held in Cracow, 10-11 July 2012}, volume~36. Springer
  New-York, 2013.

\bibitem[Dissmann et~al.(2013)Dissmann, Brechmann, Czado, Kurowicka,
  Di{\ss}mann, Brechmann, Czado, and Kurowicka]{Dissmann2013}
J.~Dissmann, Eike~Christian Brechmann, Claudia Czado, Dorota Kurowicka,
  J~Di{\ss}mann, Eike~Christian Brechmann, Claudia Czado, and Dorota Kurowicka.
\newblock {Selecting and estimating regular vine copulae and application to
  financial returns}.
\newblock \emph{Computational Statistics {\&} Data Analysis}, 59:\penalty0
  52--69, March 2013.

\bibitem[Elidan(2013)]{Elidan2013}
Gal Elidan.
\newblock {Copulas in machine learning}.
\newblock In \emph{Copulae in mathematical and quantitative finance}, pages
  39--60. Springer, 2013.

\bibitem[Geenens et~al.(2017)Geenens, Charpentier, and
  Paindaveine]{Geenens2017}
Gery Geenens, Arthur Charpentier, and Davy Paindaveine.
\newblock {Probit transformation for nonparametric kernel estimation of the
  copula density}.
\newblock \emph{Bernoulli}, 23\penalty0 (3):\penalty0 1848--1873, 2017.

\bibitem[Goodfellow et~al.(2014)Goodfellow, Pouget-Abadie, Mirza, Xu,
  Warde-Farley, Ozair, Courville, and Bengio]{goodfellow2014generative}
Ian Goodfellow, Jean Pouget-Abadie, Mehdi Mirza, Bing Xu, David Warde-Farley,
  Sherjil Ozair, Aaron Courville, and Yoshua Bengio.
\newblock Generative adversarial nets.
\newblock In \emph{NeurIPS}, pages 2672--2680, 2014.

\bibitem[Gretton et~al.(2007)Gretton, Borgwardt, Rasch, Sch{\"o}lkopf, and
  Smola]{gretton2007kernel}
Arthur Gretton, Karsten~M Borgwardt, Malte Rasch, Bernhard Sch{\"o}lkopf, and
  Alex~J Smola.
\newblock A kernel method for the two-sample-problem.
\newblock In \emph{NeurIPS}, 2007.

\bibitem[Grnarova et~al.(2018)Grnarova, Levy, Lucchi, Perraudin, Hofmann, and
  Krause]{grnarova2018evaluating}
Paulina Grnarova, Kfir~Y Levy, Aurelien Lucchi, Nathanael Perraudin, Thomas
  Hofmann, and Andreas Krause.
\newblock Evaluating gans via duality.
\newblock \emph{arXiv preprint arXiv:1811.05512}, 2018.

\bibitem[Guennebaud et~al.(2010)Guennebaud, Jacob, and Others]{eigenweb}
Ga{\"{e}}l Guennebaud, Beno{\^{i}}t Jacob, and Others.
\newblock {Eigen v3}, 2010.

\bibitem[Gulrajani et~al.(2017)Gulrajani, Ahmed, Arjovsky, Dumoulin, and
  Courville]{gulrajani2017improved}
Ishaan Gulrajani, Faruk Ahmed, Martin Arjovsky, Vincent Dumoulin, and Aaron~C
  Courville.
\newblock Improved training of wasserstein gans.
\newblock In \emph{NeurIPS}, 2017.

\bibitem[Haff(2013)]{Haff2013}
Ingrid~Hob{\ae}k Haff.
\newblock {Parameter estimation for pair-copula constructions}.
\newblock \emph{Bernoulli}, 19\penalty0 (2):\penalty0 462--491, 2013.

\bibitem[Heusel et~al.(2017)Heusel, Ramsauer, Unterthiner, Nessler, and
  Hochreiter]{heusel2017gans}
Martin Heusel, Hubert Ramsauer, Thomas Unterthiner, Bernhard Nessler, and Sepp
  Hochreiter.
\newblock Gans trained by a two time-scale update rule converge to a local nash
  equilibrium.
\newblock In \emph{NeurIPS}, 2017.

\bibitem[Higgins et~al.(2017)Higgins, Matthey, Pal, Burgess, Glorot, Botvinick,
  Mohamed, and Lerchner]{higgins2016beta}
Irina Higgins, Loic Matthey, Arka Pal, Christopher Burgess, Xavier Glorot,
  Matthew Botvinick, Shakir Mohamed, and Alexander Lerchner.
\newblock beta-vae: Learning basic visual concepts with a constrained
  variational framework.
\newblock \emph{ICLR}, 2017.

\bibitem[Hinton and Salakhutdinov(2006)]{hinton2006reducing}
Geoffrey~E Hinton and Ruslan~R Salakhutdinov.
\newblock Reducing the dimensionality of data with neural networks.
\newblock \emph{Science}, 313\penalty0 (5786):\penalty0 504--507, 2006.

\bibitem[Hinton and Zemel(1994)]{hinton1994autoencoders}
Geoffrey~E Hinton and Richard~S Zemel.
\newblock Autoencoders, minimum description length and helmholtz free energy.
\newblock In \emph{NeurIPS}, pages 3--10, 1994.

\bibitem[Hinton et~al.(2006)Hinton, Osindero, and Teh]{hinton2006fast}
Geoffrey~E Hinton, Simon Osindero, and Yee-Whye Teh.
\newblock A fast learning algorithm for deep belief nets.
\newblock \emph{Neural computation}, 18\penalty0 (7):\penalty0 1527--1554,
  2006.

\bibitem[Hyv{\"a}rinen and Oja(2000)]{hyvarinen2000independent}
Aapo Hyv{\"a}rinen and Erkki Oja.
\newblock Independent component analysis: algorithms and applications.
\newblock \emph{Neural networks}, 13\penalty0 (4-5):\penalty0 411--430, 2000.

\bibitem[Joe(1997)]{Joe97}
Harry Joe.
\newblock \emph{{Multivariate Models and Dependence Concepts}}.
\newblock Chapman {\&} Hall/CRC, 1997.

\bibitem[Joe(2014)]{joe2014dependence}
Harry Joe.
\newblock \emph{Dependence modeling with copulas}.
\newblock Chapman and Hall/CRC, 2014.

\bibitem[Killiches et~al.(2017)Killiches, Kraus, and Czado]{killiches2017}
Matthias Killiches, Daniel Kraus, and Claudia Czado.
\newblock {Model distances for vine copulas in high dimensions}.
\newblock \emph{Statistics and Computing}, pages 1--19, 2017.

\bibitem[Kingma and Welling(2014)]{kingma2013auto:vae}
Diederik~P Kingma and Max Welling.
\newblock Auto-encoding variational {Bayes}.
\newblock In \emph{ICLR}, 2014.

\bibitem[Kulkarni et~al.(2018)Kulkarni, Tagasovska, Vatter, and
  Garbinato]{kulkarni2018generative}
Vaibhav Kulkarni, Natasa Tagasovska, Thibault Vatter, and Benoit Garbinato.
\newblock Generative models for simulating mobility trajectories.
\newblock 2018.

\bibitem[Kurowicka and Joe(2010)]{Kurowicka2010}
Dorota Kurowicka and Harry Joe.
\newblock \emph{{Dependence Modeling}}.
\newblock World Scientific Publishing Company, Incorporated, 2010.
\newblock ISBN 978-981-4299-87-9.

\bibitem[LeCun and Cortes(2010)]{lecun-mnisthandwrittendigit-2010}
Yann LeCun and Corinna Cortes.
\newblock {MNIST} handwritten digit database.
\newblock 2010.

\bibitem[LeCun et~al.(1995)LeCun, Bengio, et~al.]{lecun1995convolutional}
Yann LeCun, Yoshua Bengio, et~al.
\newblock Convolutional networks for images, speech, and time series.
\newblock \emph{The handbook of brain theory and neural networks},
  3361\penalty0 (10):\penalty0 1995, 1995.

\bibitem[Li et~al.(2014)Li, Xiong, and Jiang]{Li2014}
Haoran Li, Li~Xiong, and Xiaoqian Jiang.
\newblock {Differentially Private Synthesization of Multi-Dimensional Data
  using Copula Functions}.
\newblock In \emph{Proc. of the 17th International Conference on Extending
  Database Technology}, number~c, pages 475--486, 2014.

\bibitem[Liu et~al.(2009)Liu, Lafferty, and Wasserman]{Liu2009}
Han Liu, John Lafferty, and Larry Wasserman.
\newblock {The Nonparanormal: semiparametric estimation of high dimensional
  undirected graphs}.
\newblock \emph{JMLR}, 10:\penalty0 2295--2328, 2009.

\bibitem[Liu et~al.(2015)Liu, Luo, Wang, and Tang]{liu2015faceattributes}
Ziwei Liu, Ping Luo, Xiaogang Wang, and Xiaoou Tang.
\newblock Deep learning face attributes in the wild.
\newblock In \emph{ICCV}, 2015.

\bibitem[Lopez-Paz(2016)]{Lopez-Paz2016}
David Lopez-Paz.
\newblock \emph{{From Dependence to Causation}}.
\newblock PhD thesis, University of Cambridge, 2016.

\bibitem[Lopez-Paz and Oquab(2016)]{lopez2016revisiting}
David Lopez-Paz and Maxime Oquab.
\newblock Revisiting classifier two-sample tests.
\newblock \emph{ICLR}, 2016.

\bibitem[Lopez-Paz et~al.(2013)Lopez-Paz, Hernandez-Lobato, and
  Sch{\"{o}}lkopf]{LopezPaz2013}
David Lopez-Paz, J~M Hernandez-Lobato, and Bernhard Sch{\"{o}}lkopf.
\newblock {Semi-supervised domain adaptation with copulas}.
\newblock \emph{NeurIPS}, 2013.

\bibitem[Makhzani et~al.(2016)Makhzani, Shlens, Jaitly, and
  Goodfellow]{makhzani2015adversarial}
Alireza Makhzani, Jonathon Shlens, Navdeep Jaitly, and Ian Goodfellow.
\newblock Adversarial autoencoders.
\newblock In \emph{ICLR}, 2016.

\bibitem[Metz et~al.(2016)Metz, Poole, Pfau, and
  Sohl-Dickstein]{metz2016unrolled}
Luke Metz, Ben Poole, David Pfau, and Jascha Sohl-Dickstein.
\newblock Unrolled generative adversarial networks.
\newblock \emph{ICLR}, 2016.

\bibitem[Nagler and Czado(2016)]{Nagler2016}
Thomas Nagler and Claudia Czado.
\newblock {Evading the curse of dimensionality in nonparametric density
  estimation with simplified vine copulas}.
\newblock \emph{Journal of Multivariate Analysis}, 151:\penalty0 69--89, 2016.

\bibitem[Nagler and Vatter(2017)]{vinecopulib}
Thomas Nagler and Thibault Vatter.
\newblock {vinecopulib: High Performance Algorithms for Vine Copula Modeling in
  C++}, 2017.

\bibitem[Nagler and Vatter(2018{\natexlab{a}})]{Nagler2018}
Thomas Nagler and Thibault Vatter.
\newblock \emph{kde1d: Univariate Kernel Density Estimation},
  2018{\natexlab{a}}.
\newblock R package version 0.2.1.

\bibitem[Nagler and Vatter(2018{\natexlab{b}})]{rvinecopulib}
Thomas Nagler and Thibault Vatter.
\newblock rvinecopulib: high performance algorithms for vine copula modeling,
  2018{\natexlab{b}}.

\bibitem[Nagler et~al.(2017{\natexlab{a}})Nagler, Schellhase, and
  Czado]{Nagler2017}
Thomas Nagler, Christian Schellhase, and Claudia Czado.
\newblock Nonparametric estimation of simplified vine copula models: comparison
  of methods.
\newblock \emph{Dependence Modeling}, 5\penalty0 (1):\penalty0 99--120,
  2017{\natexlab{a}}.

\bibitem[Nagler et~al.(2017{\natexlab{b}})Nagler, Schellhase, and
  Czado]{nagler2017nonparametric}
Thomas Nagler, Christian Schellhase, and Claudia Czado.
\newblock Nonparametric estimation of simplified vine copula models: comparison
  of methods.
\newblock \emph{Dependence Modeling}, 5\penalty0 (1):\penalty0 99--120,
  2017{\natexlab{b}}.

\bibitem[Nelsen(2007)]{nelsen2007introduction}
Roger~B Nelsen.
\newblock \emph{An introduction to copulas}.
\newblock Springer Science \& Business Media, 2007.

\bibitem[Netzer et~al.(2011)Netzer, Wang, Coates, Bissacco, Wu, and
  Ng]{netzer2011reading}
Yuval Netzer, Tao Wang, Adam Coates, Alessandro Bissacco, Bo~Wu, and Andrew~Y
  Ng.
\newblock Reading digits in natural images with unsupervised feature learning.
\newblock In \emph{NIPS workshop on deep learning and unsupervised feature
  learning}, volume 2011, page~5, 2011.

\bibitem[Paszke et~al.(2017)Paszke, Gross, Chintala, Chanan, Yang, DeVito, Lin,
  Desmaison, Antiga, and Lerer]{paszke2017automatic}
Adam Paszke, Sam Gross, Soumith Chintala, Gregory Chanan, Edward Yang, Zachary
  DeVito, Zeming Lin, Alban Desmaison, Luca Antiga, and Adam Lerer.
\newblock Automatic differentiation in pytorch.
\newblock 2017.

\bibitem[Patki et~al.(2016)Patki, Wedge, and Veeramachaneni]{Patki2016a}
Neha Patki, Roy Wedge, and Kalyan Veeramachaneni.
\newblock {The synthetic data vault}.
\newblock \emph{Proceedings - 3rd IEEE International Conference on Data Science
  and Advanced Analytics}, pages 399--410, 2016.

\bibitem[Prim(1957)]{prim1957shortest}
Robert~Clay Prim.
\newblock Shortest connection networks and some generalizations.
\newblock \emph{The Bell System Technical Journal}, 36\penalty0 (6):\penalty0
  1389--1401, 1957.

\bibitem[{R Core Team}(2017)]{R}
{R Core Team}.
\newblock {R: A language and environment for statistical computing}, 2017.

\bibitem[Radford et~al.(2015)Radford, Metz, and
  Chintala]{radford2015unsupervised}
Alec Radford, Luke Metz, and Soumith Chintala.
\newblock Unsupervised representation learning with deep convolutional
  generative adversarial networks.
\newblock \emph{ICLR}, 2015.

\bibitem[R{\'e}nyi(1959)]{renyi1959measures}
Alfr{\'e}d R{\'e}nyi.
\newblock On measures of dependence.
\newblock \emph{Acta mathematica hungarica}, 10\penalty0 (3-4):\penalty0
  441--451, 1959.

\bibitem[Rezende and Mohamed(2015)]{rezende2015variational}
Danilo Rezende and Shakir Mohamed.
\newblock Variational inference with normalizing flows.
\newblock In \emph{ICML}, 2015.

\bibitem[Rosenblatt(1952)]{rosenblatt1952}
Murray Rosenblatt.
\newblock Remarks on a multivariate transformation.
\newblock \emph{The annals of mathematical statistics}, 23\penalty0
  (3):\penalty0 470--472, 1952.

\bibitem[Salimans et~al.(2016)Salimans, Goodfellow, Zaremba, Cheung, Radford,
  and Chen]{salimans2016improved}
Tim Salimans, Ian Goodfellow, Wojciech Zaremba, Vicki Cheung, Alec Radford, and
  Xi~Chen.
\newblock Improved techniques for training gans.
\newblock In \emph{NeurIPS}, 2016.

\bibitem[Scaillet et~al.(2007)Scaillet, Charpentier, and
  Fermanian]{scaillet2007}
Olivier Scaillet, Arthur Charpentier, and Jean-David Fermanian.
\newblock {The estimation of copulas: Theory and practice}.
\newblock Technical report, Ensae-Crest and Katholieke Universiteit Leuven,
  NP-Paribas and Crest; HEC Geneve and Swiss Finance Institute, 2007.

\bibitem[Sch{\"{a}}ling(2011)]{Schaling2011}
Boris Sch{\"{a}}ling.
\newblock \emph{{The Boost C++ Libraries}}.
\newblock 2011.

\bibitem[Schepsmeier and St{\"{o}}ber(2014)]{Schepsmeier2014}
Ulf Schepsmeier and Jakob St{\"{o}}ber.
\newblock {Derivatives and Fisher information of bivariate copulas}.
\newblock \emph{Statistical Papers}, 55\penalty0 (2):\penalty0 525--542, May
  2014.

\bibitem[Sheather and Jones(1991)]{Sheather1991}
Simon~J Sheather and Michael~C Jones.
\newblock A reliable data-based bandwidth selection method for kernel density
  estimation.
\newblock \emph{Journal of the Royal Statistical Society. Series B
  (Methodological)}, pages 683--690, 1991.

\bibitem[Sklar(1959)]{sklar1959}
A.~Sklar.
\newblock {Fonctions de R{\'e}partition {\`a} n Dimensions et Leurs Marges}.
\newblock \emph{Publications de L'Institut de Statistique de L'Universit{\'e}
  de Paris}, 8:\penalty0 229--231, 1959.

\bibitem[St{\"{o}}ber and Czado(2012)]{stoeber2012}
Jakob St{\"{o}}ber and Claudia Czado.
\newblock {Sampling Pair Copula Constructions with Applications to Mathematical
  Finance}.
\newblock In Jan-Frederik Mai and Matthias Scherer, editors, \emph{Simulating
  Copulas: Stochastic Models, Sampling Algorithms and Applications}, Series in
  quantitative finance. World Scientific Publishing Company, Incorporated,
  2012.

\bibitem[St{\"{o}}ber and Schepsmeier(2013)]{stoeber2013}
Jakob St{\"{o}}ber and Ulf Schepsmeier.
\newblock {Estimating standard errors in regular vine copula models}.
\newblock \emph{Computational Statistics}, 28\penalty0 (6):\penalty0
  2679--2707, 2013.

\bibitem[Tagasovska et~al.(2018)Tagasovska, Vatter, and
  Chavez-Demoulin]{tagasovska2018nonparametric}
Natasa Tagasovska, Thibault Vatter, and Val{\'e}rie Chavez-Demoulin.
\newblock Nonparametric quantile-based causal discovery.
\newblock \emph{arXiv:1801.10579}, 2018.

\bibitem[Theis et~al.(2015)Theis, van~den Oord, and Bethge]{Theis2015}
Lucas Theis, A{\"{a}}ron van~den Oord, and Matthias Bethge.
\newblock {A note on the evaluation of generative models}.
\newblock In \emph{ICLR}, 2015.

\bibitem[Tolstikhin et~al.(2018)Tolstikhin, Bousquet, Gelly, and
  Schoelkopf]{tolstikhin2017wasserstein}
Ilya Tolstikhin, Olivier Bousquet, Sylvain Gelly, and Bernhard Schoelkopf.
\newblock Wasserstein auto-encoders.
\newblock \emph{ICLR}, 2018.

\bibitem[Tolstikhin et~al.(2017)Tolstikhin, Gelly, Bousquet, Simon-Gabriel, and
  Sch{\"o}lkopf]{tolstikhin2017adagan}
Ilya~O Tolstikhin, Sylvain Gelly, Olivier Bousquet, Carl-Johann Simon-Gabriel,
  and Bernhard Sch{\"o}lkopf.
\newblock Adagan: Boosting generative models.
\newblock In \emph{NeurIPS}, pages 5424--5433, 2017.

\bibitem[Tran et~al.(2015)Tran, Blei, and Airoldi]{Tran2015}
Dustin Tran, David~M Blei, and Edoardo~M Airoldi.
\newblock {Copula variational inference}.
\newblock In \emph{NeurIPS}, 2015.

\bibitem[Vincent et~al.(2008)Vincent, Larochelle, Bengio, and
  Manzagol]{vincent2008extracting}
Pascal Vincent, Hugo Larochelle, Yoshua Bengio, and Pierre-Antoine Manzagol.
\newblock Extracting and composing robust features with denoising autoencoders.
\newblock In \emph{ICML}, pages 1096--1103, 2008.

\bibitem[Xiao et~al.(2017)Xiao, Rasul, and Vollgraf]{xiao2017online}
Han Xiao, Kashif Rasul, and Roland Vollgraf.
\newblock Fashion-mnist: a novel image dataset for benchmarking machine
  learning algorithms.
\newblock 2017.

\bibitem[Xie et~al.(2016)Xie, Girshick, and Farhadi]{xie2016unsupervised}
Junyuan Xie, Ross Girshick, and Ali Farhadi.
\newblock Unsupervised deep embedding for clustering analysis.
\newblock In \emph{ICML}, 2016.

\bibitem[Xu et~al.(2018)Xu, Huang, Yuan, Guo, Sun, Wu, and
  Weinberger]{xu2018empirical}
Qiantong Xu, Gao Huang, Yang Yuan, Chuan Guo, Yu~Sun, Felix Wu, and Kilian
  Weinberger.
\newblock An empirical study on evaluation metrics of generative adversarial
  networks.
\newblock \emph{arXiv preprint arXiv:1806.07755}, 2018.

\end{thebibliography}

\newpage

\appendix 

\section*{Appendix}

\section{Introduction to (vine) copulas}

%

\subsection{Copulas} \label{sec:copula}
Recall that the components of the random vector $(X_1,\dots, X_d)$ are said to be independent if and only if its joint distribution $F$ is 
given by the product of the $d$ marginals $F_i$ for $i \in \{1, \dots, d \}$, that is
\begin{align}\label{eq:joint_indep}
  F(x_1, \dots, x_d) = \prod_{i=1}^d F_i(x_i), 
\end{align}
for any $(x_1, \dots, x_d) \in \mathcal{R}^d$.
If the random variables are absolutely continuous, then differentiating \eqref{eq:joint_indep} with respect to 
$(x_1, \dots, x_d)$ implies that a similar statement hold for the densities, that is
\begin{align}\label{eq:joint_indep_dens}
  f(x_1, \dots, x_d) = \prod_{i=1}^d f_i(x_i), 
\end{align}
where $f$ is the joint density, and $f_i$  for $i \in \{1, \dots, d \}$ are the marginal densities.

However, when the variables are dependent, this statement is no longer true. 
In this case, the celebrated Sklar's theorem (see \Cref{thm:sklar} for the precise statement) says that the 
joint distribution can be written as
\begin{align}\label{eq:joint_dep}
  F(x_1, \dots, x_d) = C(F_1(x_1), \dots, F_d(x_d)), 
\end{align}
where $C$ is a \emph{copula} that acts as a coupling mechanism between the $d$ marginals.

\begin{Definition}
  A $d$-dimensional copula \emph{copula} is a multivariate cumulative distribution
  function $C: [0,1]^d \rightarrow [0,1]$ for which all the marginal distributions are uniform.
\end{Definition}
In other words, for $d=2$, $C$ is a distribution such that $C(1,u) = C(u, 1) = u$ for any $u \in [0,1]$.
Note that the simplest copulas is arguably the independence copula, namely plugging
${C(u_1, \dots, u_d) = \prod_{i=1}^d u_i}$ into \eqref{eq:joint_dep} leads to \eqref{eq:joint_indep}.

An intuitive way to understand the copula corresponding to a given joint distribution $F$ and marginal distributions
$F_i$ for $i \in \{1, \dots, d \}$ is as the distribution of the so-called probability integral transform (PIT) of the marginals.
\begin{Definition}
  The \emph{probability integral transform} (PIT) of a random variable $X$ with distribution $F_X$ 
  is the random variable $U = F_X(X)$.
\end{Definition}
Because the PIT of any random variable is uniformly distributed\footnote{$\mathbb{P}[U \leq u] = \mathbb{P}[F_X(X) \leq u] = \mathbb{P}[X \leq F_X^{-1} (u)] = F_X(F_X^{-1}(u)) = u$}, 
the joint distribution of the vector of PITs $(U_1, \dots, U_d)$ with $U_i = F_i(X_i)$ for $i \in \{1, \dots, d \}$ is a copula, namely $C$.
A similar idea has an important consequence when one aims at sampling from the joint distribution $F$.
Because it is well known that, if $U \sim \mathcal{U}[0,1]$ and $F_X^{-1}$ is the inverse cumulative
distribution of $X$, then $F_X^{-1}(U) \sim X$, transforming samples from $C$ into samples from $F$ 
is straightforward: if $(U_1, \cdots, U_d) \sim C$, then $X_i = F_i^{-1}(U_i)$ for $i \in \{1, \dots, d \}$
implies that $(X_1, \dots, X_d) \sim F$.
While it looks like simply transforming a $d$-dimensional sampling problem into another $d$-dimensional 
sampling problem, vine copulas represent a model class for $C$ that is flexible and yet easy to sample from.

\begin{figure}
\centering
\includegraphics[width=0.9\textwidth]{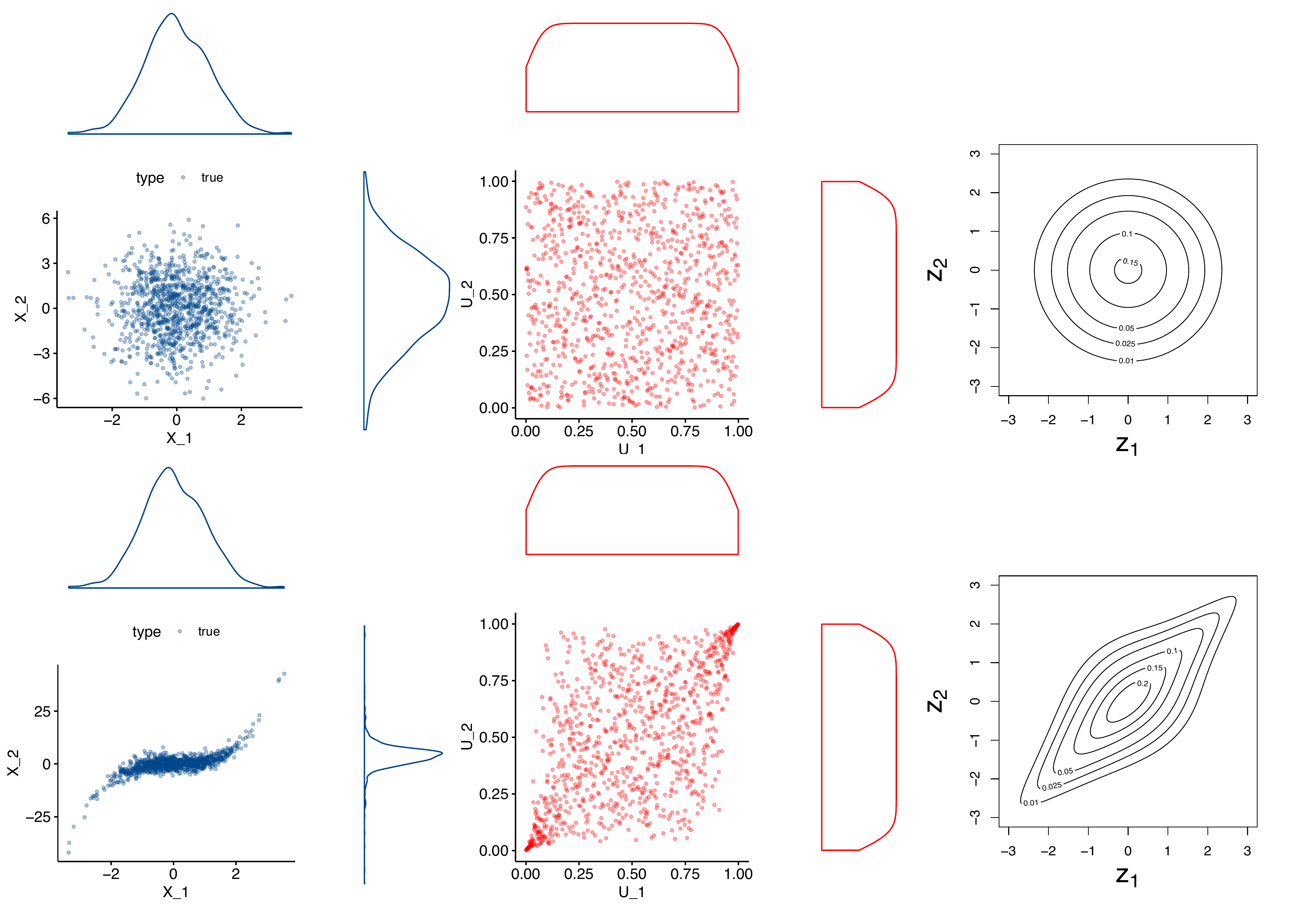}
\caption{Copula estimation. By row - (top) independent, (bottom) dependent variables. By column - (left) original data, (middle) pseudo-data after PIT, (right) estimated copula density.}
\label{fig:copula_ex}
\end{figure}

Viewing any joint distribution through this copula lens further yields a useful factorization:
differentiating \eqref{eq:joint_dep} with respect to $(x_1, \dots, x_d)$ leads to
\begin{align} \label{eq:joint_dens}
 f(x_1, \dots, x_d) &= \frac{\partial^d F(x_1, \dots, x_2)}{\partial x_1 \cdots \partial x_d}
 = \frac{\partial^d C(u_1,\dots, u_d)}{\partial u_1 \cdots \partial u_d} \prod_{i=1}^d \frac{\partial F_i(x_i)}{\partial x_i}
 =  c(u_1, \dots, u_d) \prod_{i=1}^d f_i(x_i),
\end{align}
where $c$ is the so-called \emph{copula density}, and $u_i = F_i(x_i)$  for $i \in \{1, \dots, d \}$.
Hence, we can see that the joint density factorize into a product between the marginal densities, similarly as in \eqref{eq:joint_indep_dens}, 
with the copula density, which encodes the dependence.
Taking the logarithm on both sides of \eqref{eq:joint_dens}, one obtains
\begin{align*}
  \log f(x_1, \dots, x_d) = \log c(u_1, \dots, u_d) + \sum_{i=1}^d\log f_i(x_i).
\end{align*}
In other words, the factorization implies that the joint log-likelihood is the sum 
of the marginal log-likelihoods and the copula log-likelihood.
This observation can be conveniently leveraged for estimation via a two-step procedure 
where $f_i$ is first estimated by $\widehat{f}_i$ for $i \in \{1, \dots, d \}$.
Then, pseudo-observations of the copula are recovered using the estimated PITs, 
that is $u_i \approx \widehat{F}_i(x_i)$ for $i \in \{1, \dots, d \}$, 
and $c$ is then estimated by $\widehat{c}$ using the pseudo-sample. 
This procedure is exemplified in \Cref{fig:copula_ex}.

To summarize, copulas are a tool allowing to represent any multivariate distribution 
through the individual variables' marginal behaviors as well as their inter-dependencies.
While lesser known in the machine learning community, 
copulas have been widely exploited by in other fields, from economics to quantitative finance, insurance and environmental sciences;
in particular when capturing the joint tail behavior is of high importance. 
In financial risk management for instance, so-called tail events can trigger large and simultaneous losses (or gains) on portfolios. 
Consequently, multiple parametric copula families have been studied to capture lower/upper tail dependence, or no tail dependence at all.
Similarly, other families have been developed to handle asymmetries or other dependence patterns.
But such parametric families, which usually imply that the dependence between all pairs of variables is of the same kind,
are seldom flexible enough in higher dimensions. 
Such limitations have led to the development of \emph{pair-copulas constructions} (PCCs) or  \emph{vines} - hierarchical structures which allow to flexibly model high dimensional distributions by decomposing the dependence structure into \emph{pairs of (bivariate) copulas}.

\subsection{Vines} \label{sec:vines_app} 

According to \citep{Joe97,Bedford02,Czado2010}, any copula density can be decomposed into a product of $\frac{d(d - 1)}{2}$ bivariate (conditional) copula densities.
While a decomposition is not unique, it can be organized as a graphical model, a sequence of $d - 1$ nested trees, called \emph{regular vine}, \emph{R-vine}, or simply \emph{vine}.
Denoting $T_m = (V_m, E_m)$ with $V_m$ and $E_m$ the set of nodes and edges of tree $m$ for $m = 1, \dots, d-1$, the sequence is a vine if it satisfies the following set of conditions guaranteeing that the decomposition leads to a \emph{valid joint density}:
\begin{itemize}
  \item  $T_1$ is a tree with nodes $V_1 =  \lbrace 1, \dots , d \rbrace$ and edges $E_1$.
  \item For $m \geq 2$, $T_m$ is a tree with nodes $V_m = E_{m - 1}$ and edges  $E_m$.
\item (Proximity condition) Whenever two nodes in $T_m+1$ are joined by an edge, the corresponding edges in $T_m$ must share a common node.
\end{itemize}
The corresponding tree sequence is then called the \emph{structure} of the PCC and has important implications to design efficient algorithms for the estimation and sampling of such models.

\begin{figure}[H]
\centering
\subfloat[Tree $T_1$]{
\begin{minipage}[t]{.28\linewidth}
     \centering
     \includegraphics[page=1,width=\textwidth]{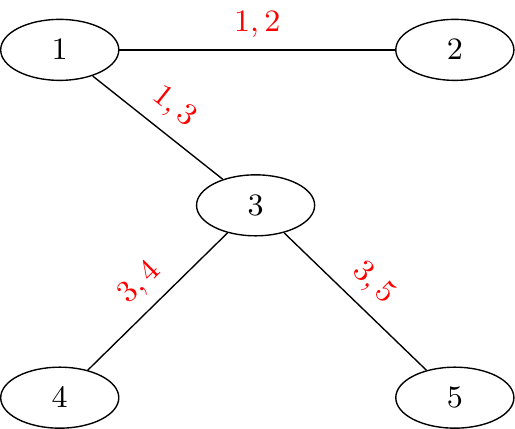}
\end{minipage}
} \label{fig:rvineex1}
%
%
\subfloat[Tree $T_2$]{
\begin{minipage}[t]{.29\linewidth}
     \centering
     \includegraphics[page=2,width=\textwidth]{tikz/rvineex.pdf}
\end{minipage} 
\label{fig:rvineex2}
} 
%
%
\subfloat[Tree $T_3$]{
\begin{minipage}[t]{.185\linewidth}
     \centering
     \includegraphics[page=3,width=\textwidth]{tikz/rvineex.pdf}
\end{minipage}
\label{fig:rvineex3}
} 
%
%
\subfloat[Tree $T_4$]{
\begin{minipage}[t]{.1\linewidth}
     \centering
     \includegraphics[page=4,width=\textwidth]{tikz/rvineex.pdf}
\end{minipage}
\label{fig:rvineex4}
} 
\caption{A vine tree sequence: the numbers represent the variables, $x,y$ the bivariate distribution of $x$ and $y$, and $x,y|z$ the bivariate distribution of $x$ and $y$ conditional on $z$. Each edge corresponds to a bivariate pair-copula in the PCC.}
	\label{sec:gampcc:RVine_fig}
\end{figure}

Each edge $e$ is associated to a bivariate copula $c_{j_e, k_e | D_e}$ (a so-called \emph{pair-copula}), with the set $D_e \in \left\{1, \cdots, d \right\}$ and the indices $j_e, k_e  \in \left\{1, \cdots, d \right\}$ forming respectively its \emph{conditioning set} and the \emph{conditioned set}. 
Finally, the joint copula density can be written as the product of all pair-copula densities
\begin{align} \label{eq:pcc_density}
  c(u_1, \cdots, u_d) = \prod_{m=1}^{d-1} \prod_{e \in E_m} c_{j_e, k_e | D_e}(u_{j_e| D_e}, u_{k_e| D_e}), 
\end{align}
where
\begin{align*}
  u_{j_e | D_e} = \mathbb{P}\left[ U_{j_e} \leq u_{j_e} \mid  \bm{U}_{D_e} = \bm{u}_{D_e} \right],
\end{align*}
and similarly for $u_{j_e | D_e}$,
with $\bm{U}_{D_e} = \bm{u}_{D_e} $ understood as component-wise equality for all components of 
$(U_1, \dots, U_d)$ and $(u_1, \dots, u_d)$ included in the conditioning set $D_e$. 
In Example~\Cref{sec:gampcc:vine_ex} we present a full example of an R vine for a 5 dimensional density.
\begin{Example}\label{sec:gampcc:vine_ex}
	The density of a PCC corresponding to the tree sequence in \Cref{sec:gampcc:RVine_fig} is
  \begin{align}
	c &= {\color{red}c_{1,2}\,  c_{1,3}\, c_{3,4}\, c_{3,5}}\, {\color{blue} c_{2,3|1}\, c_{1,4|3}\, c_{1,5|3}} {\color{green} c_{2,4|1,3}\, c_{4,5|1,3}}\, {\color{magenta} c_{2,5|1,3,4}},
	\end{align}
	where the colors correspond to the edges {\color{red} $E_1$}, {\color{blue} $E_2$}, {\color{green} $E_3$}, {\color{magenta} $E_4$}.
\end{Example}

To summarize this section, in order to construct a vine, one has to choose two components:
\begin{itemize}
  \item The structure, namely the set of trees $T_m = (V_m, E_m)$ for $m = 1, \dots, d-1$. 
  \item The pair-copulas, namely the models for $ c_{j_e, k_e | D_e}$ for $e \in E_m$ and $m = 1, \dots, d-1$.
\end{itemize}
To fix ideas, it is easier to start by assuming the structure to be known.

\subsubsection{Estimating the pair-copulas} \label{sec:vines_est}
To answer how one could estimate the pair-copulas is closely related whether one can evaluate the density in \eqref{eq:pcc_density}:
if one can evaluate the density, then taking it's logarithm and finding the MLE would be straightforward.
While it would be impractical for high-dimensional data, the factorization as a product of pair-copulas paves the 
way for a sequential procedure.
Indeed, taking the logarithm of both sides of \eqref{eq:pcc_density}, we have
\begin{align} \label{eq:pcc_loglik}
  \log c(u_1, \cdots, u_d) = \sum_{m=1}^{d-1} \sum_{e \in E_m} \log c_{j_e, k_e | D_e}(u_{j_e| D_e}, u_{k_e| D_e}). 
\end{align}
One can thus use \eqref{eq:pcc_loglik} to proceed in a tree-wise fashion, starting with $m=1$, with all pairs in a given tree, that is $e \in E_m$, being estimated in parallel.

Assuming the marginal distributions to be known, one can simply proceed with pseudo-observations $(U_1, \cdots, U_d)$ with $U_i = F_i(X_i)$
to estimate the pairs in the first tree (i.e., when $m=1$).
It works because, for those pairs, the conditioning set is empty, that is $D_e = \emptyset$.
When the marginal distributions are unknown, one can proceed similarly using $\widehat{F}_i(X_i)$.
But for the higher trees (i.e., when $m > 1$), the decomposition involves conditional distributions like $U_{j_e}|\bm{U}_{D_e}$ 
with a non-empty conditioning set, that is $D_e \neq \emptyset$.

It turns out that the arguments for pair-copulas in any tree $m>1$ can be expressed recursively using conditional distributions corresponding to bivariate copulas in
the previous tree (i.e., $m-1$) as follows. 
Let $e \in E_m$ be an edge of tree $m$ and $l_e \in D_e$ be another index such that $c_{j_e,l_e \mid D_e \setminus l_e} $ is a pair-copula in tree $m-1$, and define $D^{\prime}_e = D_e \setminus l_e$. 
Then we have that 
\begin{align*}
  u_{j_e | D_e} = h_{j_e, l_e \mid D^{\prime}_e} (u_{j_e | D'_e}, u_{l_e | D'_e}  )
\end{align*}
where the so-called $h$-function is defined as
\begin{align*}
  h_{j_e, l_e \mid D^{\prime}_e}(u_1,  u_2) := \int_0^{u_1} c_{j_e,l_e \mid D^{\prime}_e}(v, u_2)dv = \frac{\partial C_{j_e, l_e \mid D_e^{\prime}}(u_1, u_2)}{\partial u_2}.
\end{align*}
In each step of this recursion the conditioning set $D_e$ is reduced by one element, until 
we eventually reach the first tree with $D_e = \emptyset$.
Note that, in a vine, for any edge $e$, the existence of an index $l_e$ such that $c_{j_e,l_e \mid D_e \setminus l_e} $ is a pair-copula in tree $m-1$ is guaranteed.
This allows us to write any of the required conditional distributions as a recursion over $h$-functions that directly linked to the pair-copula densities in previous trees.
As such, assuming the structure to be known, a sequential algorithm to estimate the pair-copulas can be described as follow:

\begin{enumerate}
  \item Set $m = 1$ and estimate all pair-copulas for the first tree using $(U_1, \cdots, U_d)$.
  \item Set $m = m + 1$ and compute the conditional distributions $u_{j_e | D_e}$ and $u_{k_e | D_e}$ for $e \in E_m$.
  \item Estimate all pair-copulas in tree $m$ using $u_{j_e | D_e}$ and $u_{k_e | D_e}$ for $e \in E_m$.
  \item If $m=d-1$, all pairs have been estimated. Otherwise, go to step 2.
\end{enumerate}
The procedure is generic in the sense that it can be used with any bivariate copula estimator, 
and we refer to Algorithm 1 in \citep{Nagler2017} for its pseudocode.
Note that the decomposition can also be truncated by replacing the termination condition 
at step 4 using any truncation level smaller than $d - 1$.
Finally, for each pair-copula, one could also estimate different models at step 3 and select 
the best one according to some suitable criterion (e.g., AIC or BIC).
One important question that we brushed aside is: given that the structure is generally unknown, how can we also select it?

\subsubsection{Selecting the structure}  \label{sec:vines_est2}

To learn the structure for a dataset where it is unknown, multiple solutions have been proposed. 
In this paper, as it is most common in the vine literature, we use the so-called Dissmann algorithm, first proposed in \citep{Dissmann2013}:
This algorithm represents a greedy heuristic aiming at capturing higher dependencies in the lower trees.
The intuition is that higher-tree represent higher-order interactions, which are harder to estimate.
As such, one should prioritize modeling the most important patterns in lower trees.
This is achieved by finding the maximum spanning tree (MST) using a dependence measure as edge weights.
For instance, the absolute value of the empirical Kendall's $\tau$ for monotone dependencies or the maximal correlation \citep{renyi1959measures} for more general patterns are popular choices. 
To compute the MST, most implementations use Prim's algorithm \citep{prim1957shortest}.
Letting $\tau$ denote a generic bivariate dependence measure, the sequential algorithm mentioned above can thus be modified in a straightforward manner:
\begin{enumerate}
  \item Set $m = 1$ and compute the dependence $\tau(U_i, U_j)$ for all pairs $1 \leq i < j \leq d$.
   While this defines a complete graph, only keep the edges corresponding to the MST in $E_m$.
    Finally, estimate all pair-copulas for the first tree as before.
  \item Set $m = m + 1$ and compute the conditional distributions $u_{j_e | D_e}$ and $u_{k_e | D_e}$, as well as the dependence $\tau(u_{j_e | D_e}, u_{k_e | D_e})$ for all pairs where $e$ is an edge allowed by the proximity condition.
    Only keep the edges corresponding to the MST in $E_m$.
  \item Estimate all pair-copulas in tree $m$ using $u_{j_e | D_e}$ and $u_{k_e | D_e}$ for $e \in E_m$.
  \item If $m=d-1$, all pairs have been estimated. Otherwise, go to step 2.
\end{enumerate}
Note that step 2 can be implemented efficiently by observing that, while conditional distributions might appear in multiple candidate edges, they can be computed only once and stored for further use.
The resulting estimation and structure selection procedure is summarized in Algorithm 2 of \citep{Nagler2017}.

\subsection{Assumptions for the consistency and asymptotic normality of the kernel bivariate copula estimator}
\label{sec:est_assumptions}
\begin{enumerate}
\item[(B1)] $\partial_u C(u,v)$ and $\partial_{uu} C(u,v)$ exist and are continuous on $(u,v) \in(0,1) \times [0,1]$, and there exists a constant $Q_1$ such that $\vert \partial_{uu} C(u,v) \vert \leq Q_1/u (1-u)$ for $(u,v) \in(0,1) \times [0,1]$.
\item[(B2)] $\partial_v C(u,v)$ and $\partial_{vv} C(u,v)$ exist and are continuous on $(u,v) \in[0,1] \times (0,1)$, and there exists a constant $Q_2$ such that $\vert \partial_{vv} C(u,v) \vert \leq Q_2/v (1-v)$ for $(u,v) \in[0,1] \times (0,1)$.
\item[(B3)] The density $c(u,v) = \partial_{uv} C(u,v)$ admits continuous second-order partial derivatives in $(0,1)^2$ and there exists a constant $Q_0$ such that, for $(u,v) \in (0,1)^2$, ${c(u,v) \leq Q_0 \min \left(\frac{1}{u(1-u)},\frac{1}{v(1-v)} \right)}$.
\end{enumerate}
\vspace{-0.5cm}

\section{The VCAE algorithm} \label{sec:vcae_algo}

The algorithm for vine copula autoencoders is given in~\Cref{alg:vcae}.

\begin{minipage}{.7\linewidth}
\begin{algorithm}[H] 
\caption{Vine Copula Autoencoder}
\label{alg:vcae}
\begin{algorithmic}[width=0.5\textwidth]
\STATE {\bfseries \textbf{Input}:} train set $X$ of $\lbrace x_1, x_2, ... x_n \rbrace$ images.
\STATE 1. Train AE component with $X$: \\
 $f \leftarrow encoder$ \\
$g \leftarrow decoder$
\item[]

\STATE \textbf{2}. Encode train set with $f:$ \\
$\phi(X) \leftarrow f(X)$
\item[]

\STATE \textbf{3}. Fit a vine copula  $c$ using encoded features: \\
$ c \leftarrow  \lbrace \phi_1, \phi_2, ... \phi_n \rbrace$ (as described in \Cref{sec:vine_const} and \Cref{sec:estimation}).
\item[]

\STATE \textbf{4}. Sample random observations form $c$: \\
$\phi^\prime  \leftarrow c(\phi)$ (as in \Cref{sec:sampling})
\item[]

\STATE \textbf{5}. Decode the random features: \\
$X^\prime  \leftarrow g(\phi ^\prime)$
\item[]

\STATE {\bfseries \textbf{Output}:} generated images $X ^ \prime$.
\end{algorithmic}
\end{algorithm}
\end{minipage}

\subsection{Variations of VCAE}

\paragraph{Conditional VCAE}

Since the vine estimation and the AE training are independent in our approach, we can do steps 3--5 in~\Cref{alg:vcae} per class label (fit a vine per class feature) which makes the implementation of Conditional VCAE straightforward.

\paragraph{DEC-VCAE}

For the implementation of the DEC-VCAE we followed the instructions from the authors in \cite{xie2016unsupervised}. 
A difficulty with AEs is that the encoded features are typically entangled, even when the AE reconstruction is accurate. 
Therefore we enforce some clustering. 
We start with an pre-trained AE and then optimize a two-term loss function: the clustering and the reconstruction loss.

\section{Additional experiments}

\subsection{Toy datasets} \label{sec:toy}

Similarly to related generative model literature \cite{gulrajani2017improved, tolstikhin2017adagan}, we test our method on two-dimensional toy datasets. 
Since this is a 2D case, we use bivariate copulas with nonparametric marginal densities for the estimation and sampling.
The three datasets are ring of isotropic Gaussians with 8 modes, $5 \times 5$ grid of isotropic Gaussians and the swiss roll dataset. 
These datasets have proven to be challenging for GANs due to the mode collapse issues \cite{gulrajani2017improved, tolstikhin2017adagan}. 
They motivate how the flexibility of nonparametric copulas can be leveraged, and we additionally compare to a baseline Gaussian copula.
From \Cref{fig:toy_datasets}, we observe the benefits of using nonparametrics; while fitting such datasets is easy, it is clear that the Gaussian assumption is not suitable in such cases (except for the grid of Gaussians).

We further confirm this quantitatively in \Cref{tab:toy_datasets}, where we repeat the experiment on 100 random datasets of each type, and present the average and standard deviations for both copula families.
To evaluate the sampled images, additionally to the MMD, we use the negative log-likelihood (NLL) and \emph{coverage}\footnote{Coverage measures the probability mass of the true data covered by the approximate density of the learned model as $C:= \mathbb{P}_{\rm data}[d\mathbb{P}_{\rm model} > t]$ where $t$ is  selected such that $\mathbb{P}_{\rm model}[d\mathbb{P}_{\rm model} > t] = \alpha$ and where $d\mathbb{P}_{\rm model}$ denotes the model density function. 
We set $\alpha=0.95$ as in the original paper.}, a closely related metric \cite{tolstikhin2017adagan}.
As expected, nonparametrics provide better samples according to the three two-sample metrics.

\begin{table}[t]
\begin{minipage}[b]{.48\textwidth }%
\footnotesize\centering
\small
\caption{\strut Evaluation on toy datasets for nonparametric and Gaussian copula. Average and standard deviations from 100 repetitions.}

  \resizebox{6cm}{!}{
\begin{tabular}{@{}llll@{}}
\toprule
\multicolumn{1}{c}{} & \multicolumn{1}{c}{\textbf{Ring}} & \multicolumn{1}{c}{\textbf{Grid}} & \multicolumn{1}{c}{\textbf{Swiss roll}} \\ \midrule
\multicolumn{4}{c}{\textbf{nonparametric}}                                                                                             \\ \midrule
\textbf{NLL $\uparrow$}         & -2.47(0.15)                       & -3.77(0.2)                        & -5.23(0.05)                             \\
\textbf{Coverage $\uparrow$}    & 0.93(0.02)                        & 0.94(0.02)                        & 0.99(0.01)                              \\
\textbf{MMD $\downarrow$}         & 0.18(0.02)                        & 0.15(0.16)                        & 0.32(0.03)                              \\ \midrule
\multicolumn{4}{c}{\textbf{Gaussian}}                                                                                                  \\ \midrule
\textbf{NLL $\uparrow$}         & -2.98(0.05)                       & -3.34(0.07)                       & -6.21(0.05)                             \\
\textbf{Coverage $\uparrow$}    & 0.95(0.02)                        & 0.96(0.014)                       & 0.93(0.03)                              \\
\textbf{MMD $\downarrow$}         & 0.33(0.02)                        & 0.14(0.02)                        & 0.38(0.02)                              \\ \bottomrule
\end{tabular}}
\label{tab:toy_datasets}
\hspace{2em}
\medskip
\hrule height 0pt
\end{minipage}%
\begin{minipage}[b]{.45\textwidth}
\centering
\includegraphics[width=\textwidth]{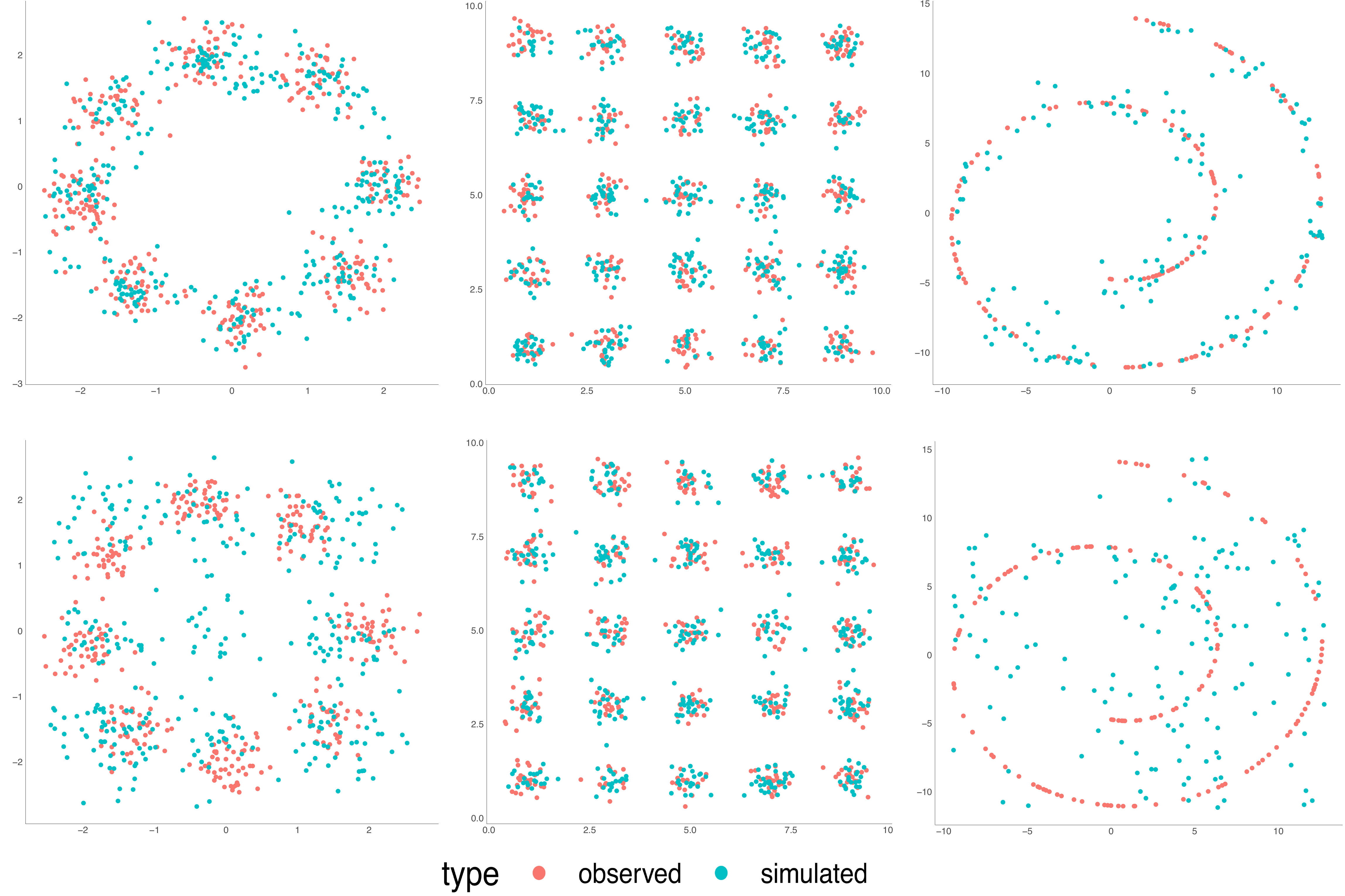}
\captionof{figure}{\strut Copula generated data - top row nonparametric, bottom row Gaussian copula.\label{fig:toy_datasets}}
\hrule height 0pt
\end{minipage}
\end{table}

\subsection{Various evaluation metrics for the MNIST and CelebA dataset} \label{sec:metrics_mnist_celeba}

\begin{figure}[h]
\centering
\includegraphics[width=0.7\textwidth]{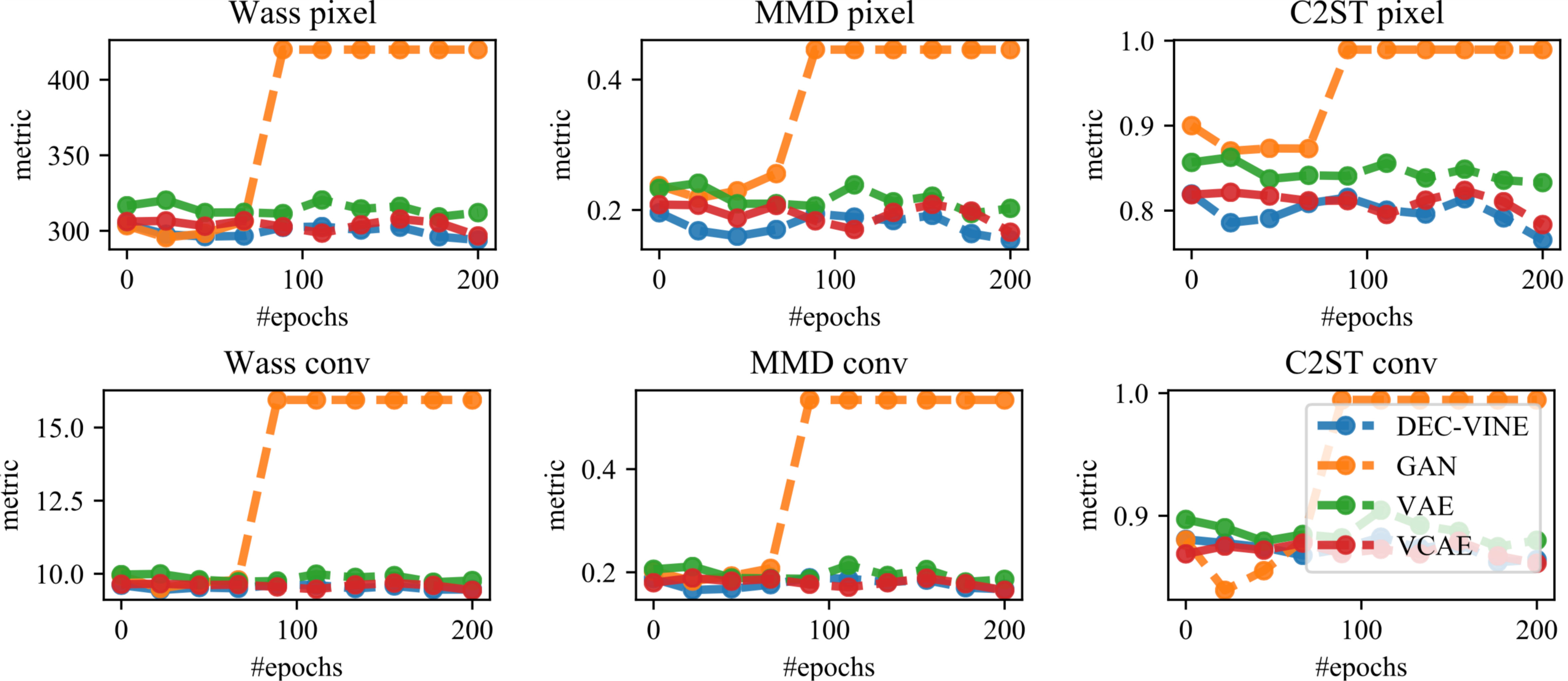}
\caption{Various evaluation scores for all baselines on the \textbf{MNIST} dataset.}
\label{fig:mnist_scores}
\end{figure}

\begin{figure}[h]
\centering
\includegraphics[width=0.9\textwidth]{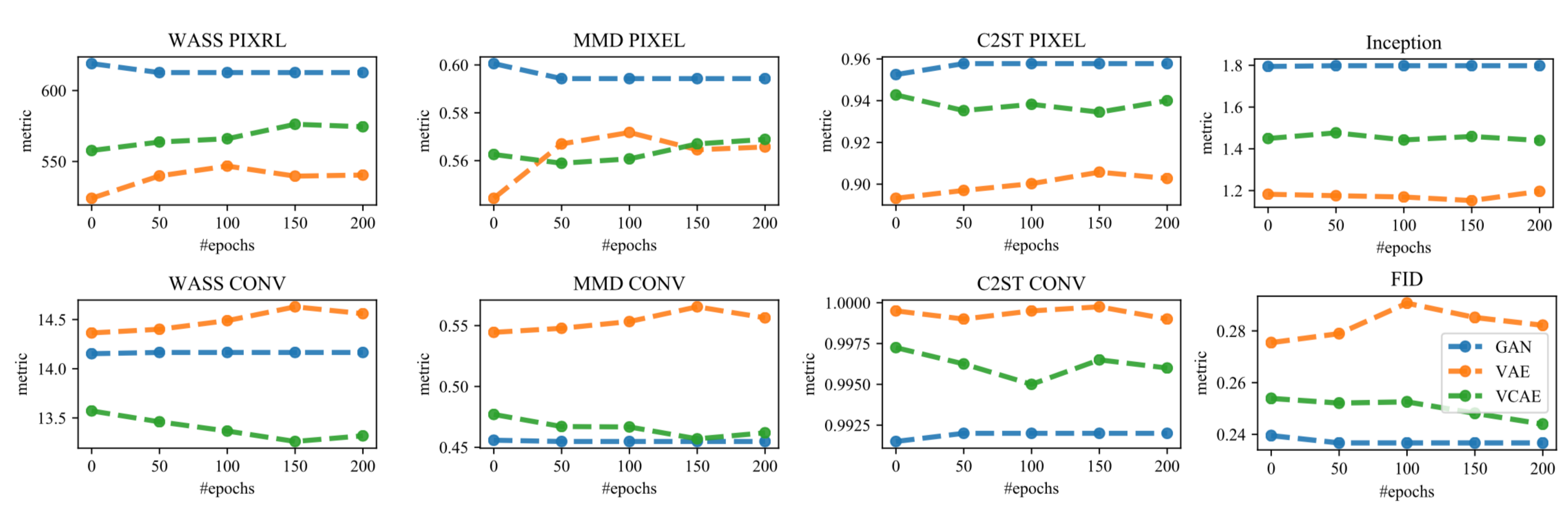}
\caption{Various evaluation scores for all baselines on the \textbf{CelebA} dataset.}
\label{fig:celeba_scores}
\end{figure}

In \Cref{fig:mnist_scores} and \Cref{fig:celeba_scores}, we present the evaluation scores for all baselines 
on the \textbf{MNIST} and \textbf{CelebA} datasets.
Note that, in the evaluation framework that we use \citep{xu2018empirical}, the Inception Score and FID are based on ImageNet features.
Therefore those scores are not suitable for binary images and excluded from \Cref{fig:mnist_scores}.

The results in \Cref{fig:celeba_scores} show that DCGAN has a slight advantage over VAE and VCAE when methods are evaluated in feature space, while VCAE outperforms VAE on all metrics. 
In this experiment we used adaptive learning rate for DCGAN \footnote{reducing the learning rate by 10 after 30th epoch} to evaluate the scores on more than 30 epochs.

\subsection{Interpolation in latent space} \label{sec:interpolation}
\begin{figure}[h]
\centering
\includegraphics[width=0.8\textwidth]{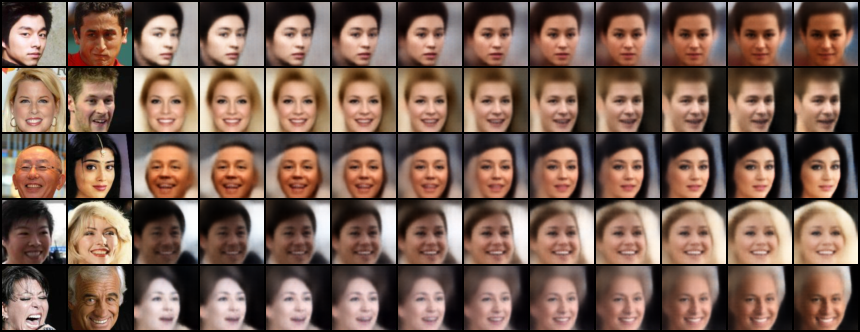}
\caption{Interpolation in latent space between two real samples (shown in the first two columns) with a VCAE trained on \textbf{CelebA}}
\label{fig:interpoation_celeb}
\end{figure}

\Cref{fig:interpoation_celeb} shows that the transitions for VCAE are smooth and without any sharp changes or unexpected samples in-between when walking the latent space by linear interpolation between two test samples as in \cite{radford2015unsupervised}.
This is not explicitly related to VCAE generative models since we do not train an end-to-end model, however it is important to show that the AE network we use did not simply memorize images.

\subsection{The trade-off between time complexity and sample quality}\label{sec:complexity_quality}

\begin{figure}[H]
\centering
\includegraphics[width = \textwidth]{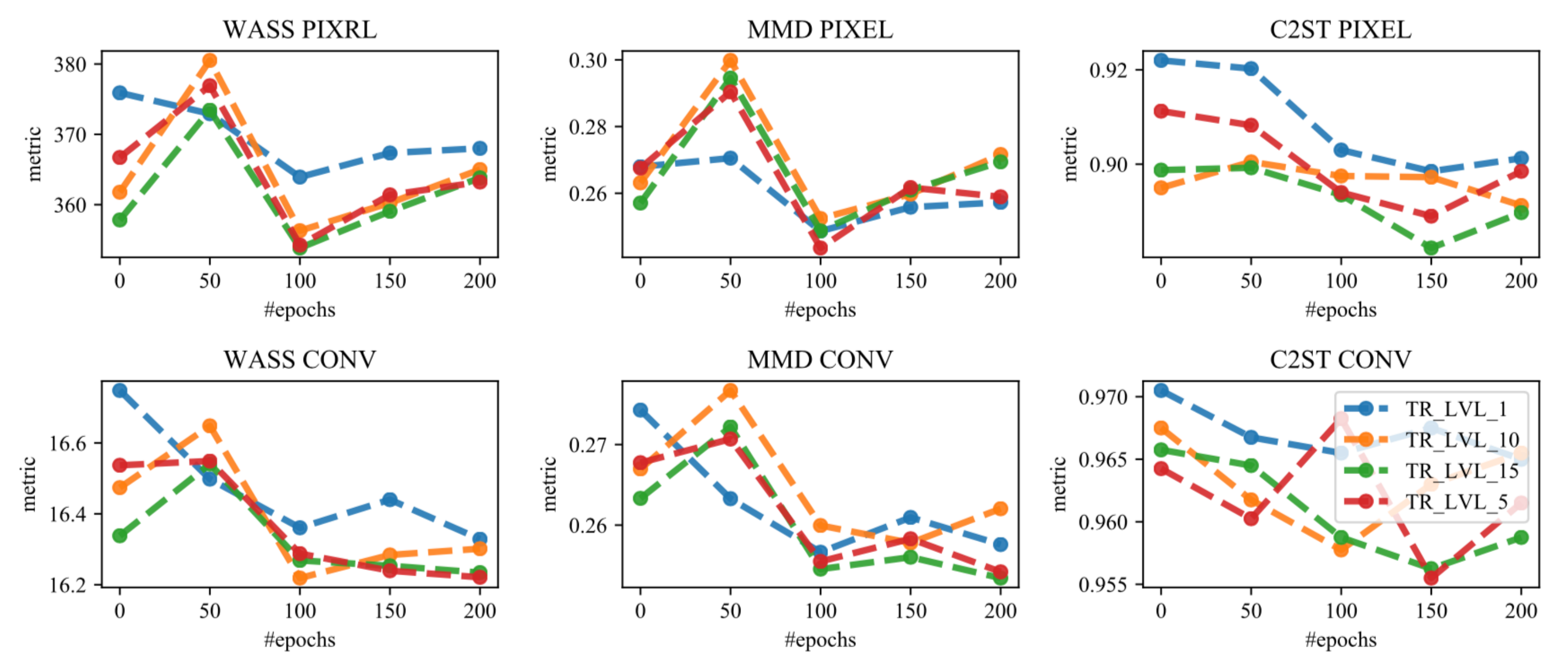}
\caption{Quantitative evaluation of various truncation levels for the VCAE on \textbf{FashionMNIST}.}
\label{fig:fashion_metrics}
\end{figure}

\begin{figure}[H]
\centering
\includegraphics[width = \textwidth]{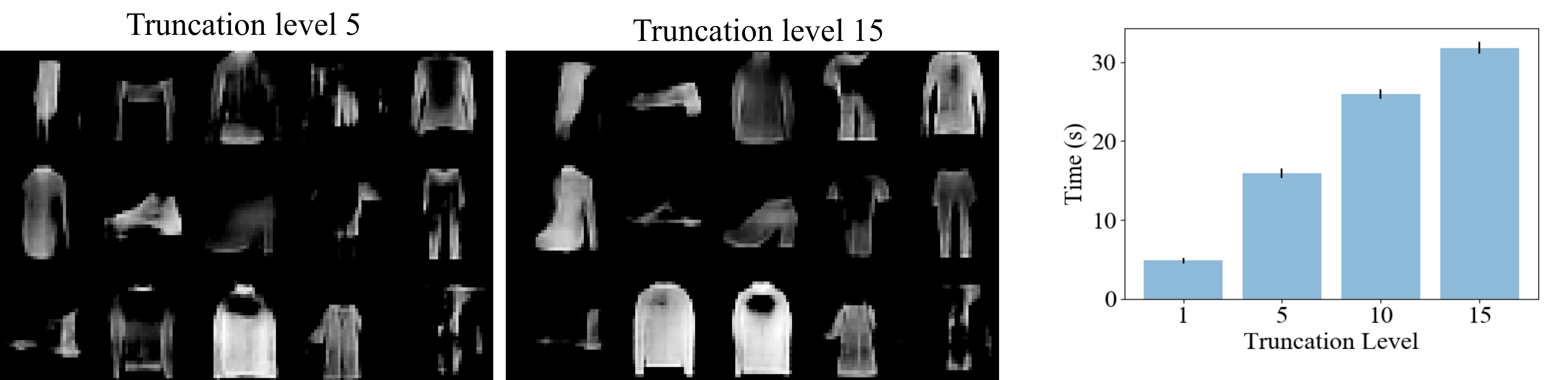}
\caption{Qualitative evaluation of various truncation levels (left panel) and computation time with respect to the vine depth (right panel) for the VCAE on \textbf{FashionMNIST}.}
\label{fig:fashion_trunc}
\end{figure}

To explore the effect of the choice for truncation level, i.e. the depth of the vine (number of trees) over the quality of the produced VCAE samples, we include an ablation study on the FashionMNIST dataset \citep{xiao2017online}.
The quantitative and qualitative evaluation in \Cref{fig:fashion_metrics} and \Cref{fig:fashion_trunc} suggest that higher level of truncation provide better samples, at the expected cost of longer computation times.

\section{Additional details on the experiments} \label{sec:architectures}

We use the same AE architecture for VCAE, DEC-VCAE and VAE as described below. 
All the AEs were trained by minimizing the Binary Cross Entropy Loss.

\subsection{MNIST}

The only transformation performed on this dataset is a padding of 2. 
By doing so we are able to use the same architecture for multiple datasets.
We use CNNs for the encoder and the decoder whose architectures are as follows:

\begin{itemize}
\item Encoder:

\begin{align*}
x \in R^{32 \times 32}  \rightarrow Conv_{32} \rightarrow BN \rightarrow ReLU \\
\rightarrow Conv_{64} \rightarrow BN \rightarrow ReLU \\
\rightarrow Conv_{128} \rightarrow BN \rightarrow ReLU \\
\rightarrow FC_{10}
\end{align*}

\item Decoder:

\begin{align*}
z \in R^{10}  \rightarrow FC_{100} \rightarrow ConvT_{128} \rightarrow BN \rightarrow ReLU \\
\rightarrow ConvT_{64} \rightarrow BN \rightarrow ReLU \\
\rightarrow ConvT_{128} \rightarrow BN \rightarrow ReLU \\
\rightarrow FC_{1}
\end{align*}

\item DCGAN Generator:
\begin{align*}
z \in R^{100} \rightarrow \rightarrow ConvT_{1} \rightarrow BN \rightarrow ReLU \\
\rightarrow ConvT_{128} \rightarrow BN \rightarrow ReLU \\
\rightarrow ConvT_{64} \rightarrow BN \rightarrow ReLU \\
\rightarrow ConvT_{32} \rightarrow BN \rightarrow ReLU \\
\rightarrow ConvT_{16} \rightarrow BN \rightarrow ReLU \\
\rightarrow Tanh_{1}
\end{align*}

\item DCGAN Discriminator:
\begin{align*}
 Conv_{1} \rightarrow BN \rightarrow LeakyReLU \\
 \rightarrow Conv_{16} \rightarrow BN \rightarrow LeakyReLU \\
\rightarrow Conv_{32} \rightarrow BN \rightarrow LeakyReLU \\
\rightarrow Conv_{64} \rightarrow BN \rightarrow LeakyReLU \\
\rightarrow Conv_{128} \rightarrow BN \rightarrow LeakyReLU \\
\rightarrow Sigmoid_{1}
\end{align*}

\end{itemize}
with all (de)convolutional layers have 4 $\times$ 4 filters, a stride of 2, and a padding of 1. 
We use BN to denote batch normalization and ReLU for rectified linear units and FC for fully connected layers. 
We denote $Conv_k$ the convolution with $k$ filters.
Leaky ReLU was used with negative slope = 0.2 everywhere.

\subsection{SVHN }

For SVHN we use the data as is without any preprocessing. The architectures are:

\begin{itemize}
\item  Encoder:
\begin{align*}
x \in R^{3 \times 32 \times 32}  \rightarrow Conv_{64} \rightarrow BN \rightarrow LeakyReLU \\
\rightarrow Conv_{128} \rightarrow BN \rightarrow LeakyReLU \\
\rightarrow Conv_{256} \rightarrow BN \rightarrow LeakyReLU \\
\rightarrow FC_{100} \rightarrow   FC_{20}
\end{align*}

\item Decoder:
\begin{align*}
z \in R^{20}  \rightarrow FC_{100} 
\rightarrow ConvT_{256} \rightarrow BN \rightarrow ReLU \\
\rightarrow ConvT_{128} \rightarrow BN \rightarrow ReLU \\
\rightarrow ConvT_{64} \rightarrow BN \rightarrow ReLU \\
\rightarrow ConvT_{32} \rightarrow BN \rightarrow ReLU \\
 \rightarrow FC_{1}
\end{align*}

\item DCGAN Generator:
\begin{align*}
z \in R^{100} \rightarrow ConvT_{256} \rightarrow BN \rightarrow ReLU \\
\rightarrow ConvT_{128} \rightarrow BN \rightarrow ReLU \\
\rightarrow ConvT_{64} \rightarrow BN \rightarrow ReLU \\
\rightarrow ConvT_{32} \rightarrow BN \rightarrow ReLU \\
\rightarrow ConvT_{3} \rightarrow BN \rightarrow ReLU \\
\rightarrow Tanh_{1}
\end{align*}

\item DCGAN Discriminator:
\begin{align*}
 Conv_{3} \rightarrow BN \rightarrow LeakyReLU \\
 \rightarrow Conv_{32} \rightarrow BN \rightarrow LeakyReLU \\
\rightarrow Conv_{64} \rightarrow BN \rightarrow LeakyReLU \\
\rightarrow Conv_{128} \rightarrow BN \rightarrow LeakyReLU \\
\rightarrow Conv_{256} \rightarrow BN \rightarrow LeakyReLU \\
\rightarrow Sigmoid_{1}
\end{align*}

\end{itemize}
where all (de)convolutional the layers have 4 $\times$ 4 filters, a stride of 2, and a padding of 1.
The rest of the notations are the same as before.

\subsection{CelebA}

For CelebA we first took central crops of 140 $\times$ 140 and then resized to resolution 64 $\times$ 64. 
Note that only Fig.~9 in the main text is not a result of this preprocessing. 
The architectures used are as follows:

\begin{itemize}
\item Encoder:
\begin{align*}
x \in R^{3 \times 64 \times 64}  \rightarrow Conv_{64} \rightarrow BN \rightarrow LeakyReLU \\
\rightarrow Conv_{128} \rightarrow BN \rightarrow LeakyReLU \\
\rightarrow Conv_{256} \rightarrow BN \rightarrow LeakyReLU  \\
\rightarrow Conv_{512} \rightarrow BN \rightarrow LeakyReLU \\
\rightarrow FC_{100} \rightarrow   FC_{100}
\end{align*}

\item Decoder:
\begin{align*}
z \in R^{100}  \rightarrow FC_{100} 
\rightarrow ConvT_{512} \rightarrow BN \rightarrow ReLU \\
\rightarrow ConvT_{256} \rightarrow BN \rightarrow ReLU \\
\rightarrow ConvT_{128} \rightarrow BN \rightarrow ReLU \\
\rightarrow ConvT_{64} \rightarrow BN \rightarrow ReLU \\
\rightarrow ConvT_{32} \rightarrow BN \rightarrow ReLU \\
 \rightarrow FC_{1}
\end{align*}

\item DCGAN Generator:
\begin{align*}
z \in R^{100} \rightarrow ConvT_{512} \rightarrow BN \rightarrow ReLU \\
\rightarrow ConvT_{256} \rightarrow BN \rightarrow ReLU \\
\rightarrow ConvT_{128} \rightarrow BN \rightarrow ReLU \\
\rightarrow ConvT_{64} \rightarrow BN \rightarrow ReLU \\
\rightarrow ConvT_{3} \rightarrow BN \rightarrow ReLU \\
\rightarrow Tanh_{1}
\end{align*}

\item DCGAN Discriminator:
\begin{align*}
 Conv_{3} \rightarrow BN \rightarrow LeakyReLU \\
\rightarrow Conv_{64} \rightarrow BN \rightarrow LeakyReLU \\
\rightarrow Conv_{128} \rightarrow BN \rightarrow LeakyReLU \\
\rightarrow Conv_{256} \rightarrow BN \rightarrow LeakyReLU \\
\rightarrow Conv_{512} \rightarrow BN \rightarrow LeakyReLU \\
\rightarrow Sigmoid_{1}
\end{align*}

\end{itemize}
where all the (de)convolutional layers have 4 $\times$ 4 filters, a stride of 2, and a padding of 1. 
Padding was set to 0 only for the last convoluitional layer of the encoder and the first layer of the decoder. 
The rest of the notations are the same as before.

\section{Code} \label{sec:code}
Our code is available at the following link: \href{https://github.com/tagas/vcae}{https://github.com/tagas/vcae}.

\section{Simulating Mobility Trajectories with copulas} \label{sec:mobility}
In related work \cite{kulkarni2018generative}, we have also compared to adversarial and recurrent based methods for sampling \emph{sequential data} (artificial mobility trajectories). 
We evaluate the generated trajectories with respect to their geographic and semantic similarity, circadian rhythms, long-range dependencies, training and generation time.
We also include two sample tests to assess statistical similarity between the observed and simulated distributions, and we analyze the privacy trade-offs with respect to membership inference and location-sequence attacks.
The results show that copulas surpass all baselines in terms of MMD score and training + simulation time. 
For more details please see \cite{kulkarni2018generative}.

\end{document}